\documentclass[11pt]{article}

\usepackage[preprint]{acl}

\usepackage{times}
\usepackage{latexsym}

\usepackage[T1]{fontenc}

\usepackage[utf8]{inputenc}

\usepackage{microtype}

\usepackage{inconsolata}

\usepackage{graphicx}
\usepackage{textcomp}
\usepackage{bbm}
\usepackage{amsmath,amssymb}
\usepackage{multirow} 
\usepackage{booktabs}
\usepackage{diagbox}
\usepackage{tikz}
\usepackage{fontawesome5}
\usepackage[table]{xcolor}  
\usepackage{tcolorbox}
\usepackage{subfigure}  
\usepackage{pifont}

\definecolor{darkgreen}{RGB}{0,100,0}  
\newcommand{\OurMODEL}{TKG-Thinker}

\title{TKG-Thinker: Towards Dynamic Reasoning over Temporal Knowledge Graphs via Agentic Reinforcement Learning}

\newcommand*{\affaddr}[1]{#1} 
\newcommand*{\affmark}[1][*]{\textsuperscript{#1}} 
\newcommand*{\email}[1]{\texttt{#1}} 

\author{
Zihao Jiang\affmark[1]$^{*}$,
Miao Peng\affmark[2]\thanks{~~Equal contribution.},
Zhenyan Shan\affmark[3],
Wenjie Xu\affmark[1],
Ben Liu\affmark[1],
Gong Chen\affmark[1],
Ziqi Gao\affmark[4],
Min Peng\affmark[3]\\
\affaddr{\affmark[1]School of Computer Science, Wuhan University, China}\\
\affaddr{\affmark[2]The Hong Kong University of Science and Technology (Guangzhou)}\\
\affaddr{\affmark[3]School of Artificial Intelligence, Wuhan University}\\
\affaddr{\affmark[4]Tsinghua Shenzhen International Graduate School, Tsinghua University}\\
\email{\{jiangzihao,bbcavendish,vingerxu,liuben123,chengongcg,pengm\}@whu.edu.cn}\\
\email{mpeng885@connect.hkust-gz.edu.cn}\\
\email{ziqigao@sz.tsinghua.edu.cn}
}

\begin{document}
\maketitle
\begin{abstract}

Temporal knowledge graph question answering (TKGQA) aims to answer time-sensitive questions by leveraging temporal knowledge bases. While Large Language Models (LLMs) demonstrate significant potential in TKGQA, current prompting strategies constrain their efficacy in two primary ways. First, they are prone to reasoning hallucinations under complex temporal constraints. Second, static prompting limits model autonomy and generalization, as it lacks optimization through dynamic interaction with temporal knowledge graph (TKG) environments. To address these limitations, we propose \textbf{\OurMODEL{}}, a novel agent equipped with autonomous planning and adaptive retrieval capabilities for reasoning over TKGs. Specifically, \OurMODEL{} performs in-depth temporal reasoning through dynamic multi-turn interactions with TKGs via a dual-training strategy. We first apply Supervised Fine-Tuning (SFT) with chain of thought data to instill core planning capabilities, followed by a Reinforcement Learning (RL) stage that leverages multi-dimensional rewards to refine reasoning policies under intricate temporal constraints. Experimental results on benchmark datasets with three open-source LLMs show that \OurMODEL{} achieves state-of-the-art performance and exhibits strong generalization across complex TKGQA settings.

\end{abstract}

\section{Introduction}

Temporal knowledge graphs (TKGs) organize factual knowledge over time and serve as an essential foundation for a wide range of knowledge-driven applications, such as recommendation systems~\cite{DBLP:conf/www/LiZLCRKL25,DBLP:journals/tkde/ChenHZWLCW25} and question answering~\cite{DBLP:conf/acl/LiuL0W0W0025,DBLP:conf/acl/GaoQKWH024}. In TKGs, facts are represented as quadruples (\textit{subject}, \textit{relation}, \textit{object}, \textit{timestamp}). Building upon this representation, temporal knowledge graph question answering (TKGQA) focuses on answering time-sensitive questions by leveraging the knowledge stored in TKGs. For instance, the question \emph{``Which team did Luka Dončić play for on 2025-02-03?''} can be answered by the quadruple \emph{(Luka Dončić, play for, Los Angeles Lakers, 2025-02-03)}.

\begin{figure}[!t]
    \centering
    \includegraphics[width=\linewidth]{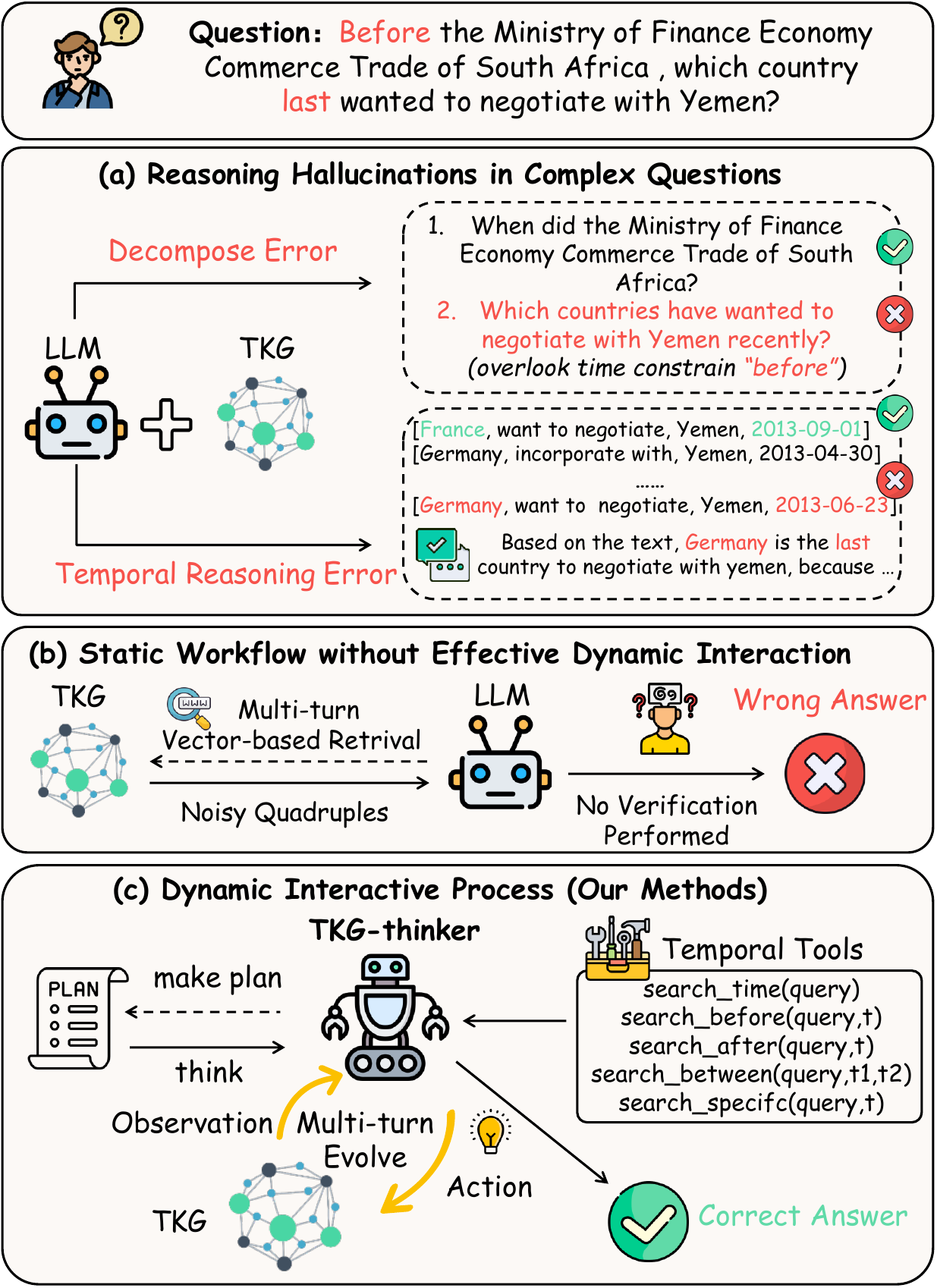}
    \caption{Comparison between \textbf{\OurMODEL{}} and existing LLM-based methods. \OurMODEL{} employs a think–action–observation loop for autonomous interaction with TKGs, enabling verified temporal reasoning.}
    \label{fig:introduction}
\end{figure}

Recently, large language models (LLMs) have demonstrated remarkable performance in tackling complex tasks~\cite{DBLP:journals/corr/abs-2412-19437,DBLP:journals/corr/abs-2505-09388,DBLP:conf/kdd/Peng0S025}. Building on this success, recent studies have increasingly focused on exploring the potential of LLMs for addressing TKGQA. For instance, some methods~\cite{TempAgent, ARI} employ few-shot prompting to guide LLM-based agents in performing temporal reasoning over TKGs, while others~\cite{gong-etal-2025-rtqa,PoK,TimeR4} decompose temporal questions into a sequence of sub-questions and combine retrieval-augmented generation (RAG) mechanisms to support step-by-step reasoning. Despite substantial progress, both paradigms struggle in complex settings for two main reasons.  

First, existing LLM-based methods are \textbf{prone to reasoning hallucinations when handling complex temporal constraints in TKGs,} including incorrect sub-question decomposition and insensitivity to fine-grained temporal constraints. As illustrated in Figure~\ref{fig:introduction}(a), when faced with the question \textit{“Before the Ministry of Finance Economy Commerce Trade of South Africa, which country last wanted to negotiate with Yemen?”}, LLMs often fail to account for critical temporal constraints such as \textit{before} and \textit{last}. As a result,  even when the relevant evidence is explicitly available in the temporal context, these methods tend to generate logically inconsistent reasoning steps and ultimately produce incorrect answers. Second, current methods suffer from \textbf{limited autonomy and suboptimal generalization} due to their reliance on static, manually-engineered workflows. More fundamentally, these models lack optimization through dynamic interaction with TKG environments, hindering the development of grounded temporal reasoning. As shown in Figure~\ref{fig:introduction}(b), this limitation leads to two critical failures: (1) retrievers often provide context that violates temporal constraints, and (2) the lack of internal verification mechanisms prevents LLMs from autonomously detecting and correcting such misaligned evidence during inference.

Regarding the limitations of static workflows, Reinforcement Learning (RL)~\cite{DBLP:journals/corr/abs-2508-20722,DBLP:journals/corr/abs-2504-13837} provides a promising paradigm for shifting to autonomous optimization. By utilizing dynamic reward signals~\cite{DBLP:journals/corr/abs-2402-03300}, RL allows LLMs to acquire complex reasoning behaviors, such as self-correction and strategic search, which are essential for navigating temporal environments~\cite{DBLP:journals/corr/abs-2503-09516}. Such emergent capabilities provide a robust mechanism to bridge the gap between static retrieval and temporally-grounded reasoning over TKGs, thereby facilitating the mitigation of hallucinations and limited exploration. Motivated by this perspective, we pose the following research question to guide our study:
\textit{Can LLMs be effectively trained to autonomously perform dynamic reasoning and retrieval in complex TKGQA scenarios via RL-based optimization?}

To address these challenges, we propose \textbf{\OurMODEL{}}, a novel agent that reformulates TKGQA as a multi-step interactive process within a dynamic environment. \OurMODEL{} performs principled question decomposition and temporal analysis through a two-stage optimization. Specifically, we first employ supervised fine-tuning on a customized dataset with chain-of-thought reasoning paths to equip the model with planning and ReAct-style capabilities~\cite{ReAct}. This stage establishes the fundamental "think–action–observation" loop and effectively alleviates the cold-start problem for subsequent optimization. In the second stage, we optimize \OurMODEL{} via RL, formalizing temporal reasoning as a sequential decision-making process. To ensure effective policy optimization, we implement a structured interaction protocol where the agent must provide explicit reasoning steps before executing predefined temporal actions (e.g., planning, time-aware retrieval), ensuring that trajectories are fully observable. Specifically, we employ a multi-objective reward mechanism that incorporates an outcome reward for factual correctness, a format reward for structured reasoning, and a retrieval reward for information coverage. This scheme enables \OurMODEL{} to internalize autonomous and dynamic reasoning behaviors in complex TKGQA scenarios. In summary, the contributions of this paper are as follows:

\begin{itemize}

    \item We introduce \textbf{\OurMODEL{}}, a novel agent capable of autonomously performing dynamic, multi-step temporal reasoning.
    
    \item To the best of our knowledge, this is the first work to explore modeling TKGQA as an RL-driven interleaved decision-making process with a time-aware interaction protocol and multi-dimensional reward design.

    \item Extensive experiments on benchmark datasets with three open-source LLMs demonstrate significant improvements over state-of-the-art TKGQA methods across multiple metrics.

\end{itemize}

\section{Related Work}

\subsection{TKGQA}

Temporal Knowledge Graph Question Answering is a challenging task that requires models to jointly reason over entities and temporal information in TKGs. Early approaches, such as: MultiQA~\cite{MULTITQ}, TempoQR~\cite{DBLP:conf/aaai/MavromatisSIAHG22}, and TSQA~\cite{DBLP:conf/acl/ShangW0022}, typically formulate TKGQA as a temporal knowledge graph completion task, relying on scoring functions to assess the plausibility of candidate facts.  Recent LLM-based methods mainly treat the question as a query over the TKGs and use retrieved evidence for reasoning. Specifically, ARI~\cite{ARI} enhances the temporal adaptability of LLMs through time-aware training and reasoning signals, while TempAgent~\cite{TempAgent} treats the LLM as an agent that performs interaction. TimeR$^{4}$~\cite{TimeR4} and PoK~\cite{PoK} strengthen LLM reasoning by improving the retrieval component, whereas RTQA~\cite{gong-etal-2025-rtqa} decomposes questions into sub-problems solved in a bottom-up manner with LLMs and TKGs. Nevertheless, these LLM-based methods still rely on manually crafted prompts, which limits their ability to autonomously detect and correct evidence.

\subsection{Agentic Reasoning with RL}
While prompting-based methods facilitate search capabilities in a training-free manner~\cite{Search_o1}, the landscape is shifting toward training-centric agentic reasoning. Advanced approaches have demonstrated that Reinforcement Learning with Verifiable Rewards (RLVR) can unlock superior reasoning abilities in LLMs~\cite{GRPO, DAPO}. This has catalyzed efforts to optimize agentic workflows—such as multi-step search~\cite{DBLP:journals/corr/abs-2503-09516, tongyi_deepresearch} and external tool integration~\cite{DBLP:journals/corr/abs-2508-20722}—using RL. Despite progress in long-horizon tasks through self-reflection~\cite{Search_and_Refine} and structured memory~\cite{Memory_R1}, these methods primarily focus on logical or mathematical tasks, but lack specialized mechanisms to navigate the intricate temporal constraints inherent in TKGs.

\section{Preliminary}
To enhance the reasoning capabilities of LLMs for TKGQA, we employ the RLVR framework, which optimizes the policy $\pi_\theta$ using deterministic rewards $r$ (e.g., execution results or exact-match accuracy). Formally, the objective is to maximize:
\begin{equation*}
\begin{aligned}
\mathcal{J}(\theta) & = \hat{\mathbb{E}}_{Q, \mathbf{y} \sim \pi_{old}} \left[ \frac{1}{G} \sum_{i=1}^G f_\epsilon (\rho_i(\theta), \hat{A}_i) \right] \\ 
& - \beta \cdot \hat{\mathbb{E}}_{Q} \left[ \mathbb{D}_{KL}[\pi_\theta(\cdot | Q) || \pi_{\text{ref}}(\cdot | Q)] \right],
\end{aligned}
\end{equation*}
where $G$ denotes the number of sampled trajectories per prompt ($G > 1$ for GRPO). $\rho_i(\theta) = \frac{\pi_\theta(y_i | Q)}{\pi_{old}(y_i | Q)}$ represents the importance sampling ratio. $f_\epsilon$ denotes the clipping function used in PPO/GRPO to stabilize updates, while $\hat{A}_i$ represents the advantage of trajectory $y_i$ computed based on verifiable rewards. The KL divergence term, scaled by $\beta$, penalizes deviations from a reference policy $\pi_{\text{ref}}$ to prevent model collapse.

\section{Methodology}

\begin{figure*}[!t]
    \centering
    \includegraphics[width=\textwidth]{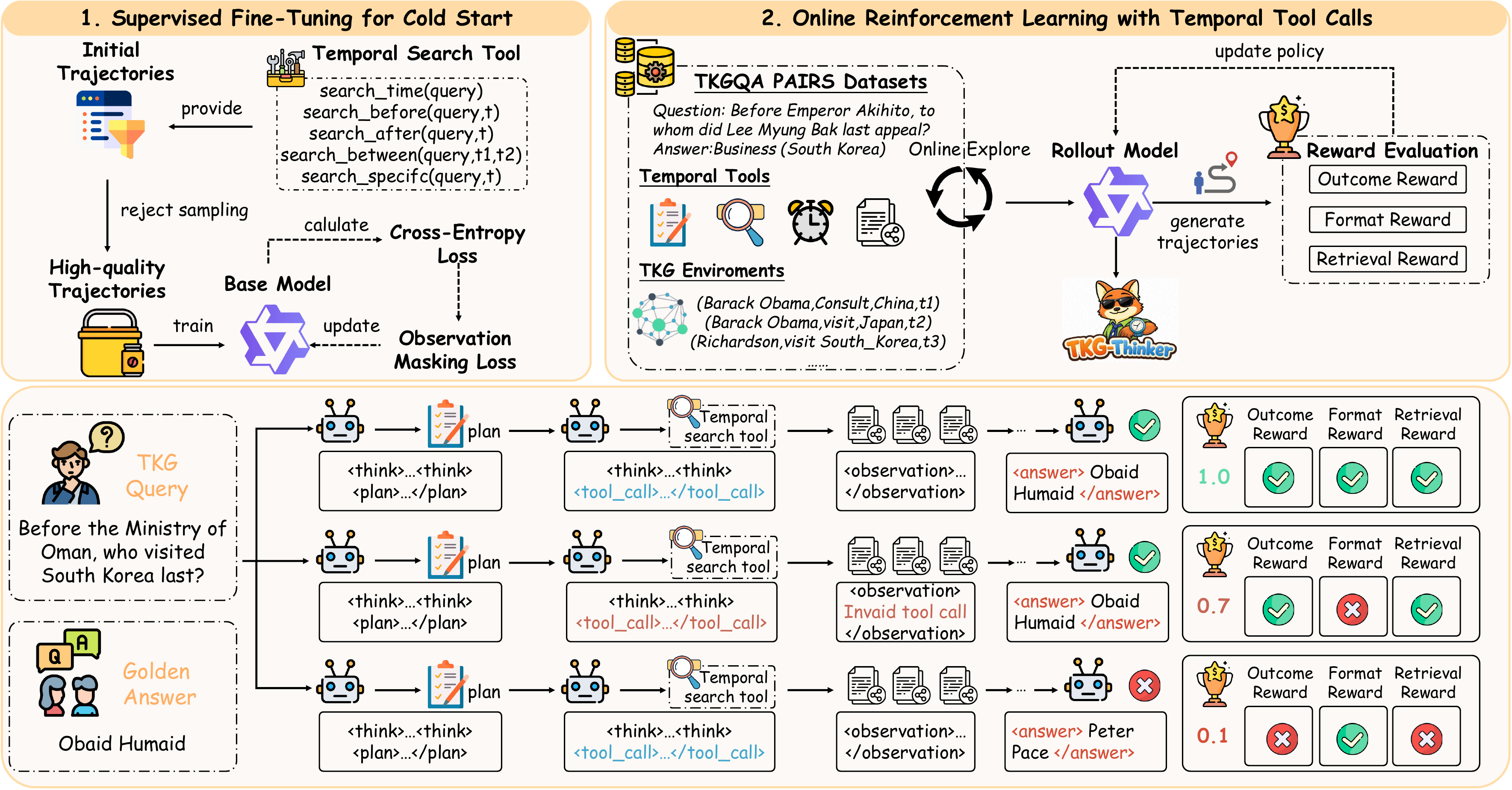}
    \caption{The overview of our proposed \OurMODEL{}. We first apply supervised fine-tuning on high-quality trajectories to mitigate the cold-start problem, and further refine the model via online reinforcement learning with temporal tool calls. The bottom panel illustrates three rollouts: a complete success, a partial success, and a failure.}
    \label{fig:Framework}
\end{figure*}

In this section, we introduce \OurMODEL{}, a novel agent equipped with autonomous planning and adaptive retrieval capabilities for reasoning over TKGs. As illustrated in Figure~\ref{fig:Framework}, \OurMODEL{} is trained through two complementary stages:
(1) Supervised Fine-Tuning (SFT) for cold start (§~\ref{Supervised Fine-Tuning for Cold Start}), and (2) Online Reinforcement Learning with Temporal Tool Calls (§~\ref{Online RL with Temporal Tool Calls}).

\subsection{Supervised Fine-Tuning for Cold Start}
\label{Supervised Fine-Tuning for Cold Start}

\OurMODEL{} relies on meaningful exploration over tool-augmented trajectories, but a generic base model does not yet know how to plan or invoke tools in the expected format, leading to low-quality rollouts~\cite{shao2025dr}. To address this, we perform SFT on trajectories generated by a strong teacher model (e.g., GPT-4o) acting as a tool-augmented agent~\cite{DBLP:journals/corr/abs-2508-13167}, thereby initializing \OurMODEL{} with a reasonable search and citation strategy before online RL. To enable the teacher model to produce tool-augmented CoT trajectories suitable for SFT, we carefully construct a prompting pipeline. Specifically, we first adopt a few-shot prompting strategy to elicit structured CoT trajectories from the teacher model, as detailed in Appendix~\ref{Few-shot Prompt for Initial Trajectory Generation}. However, not all generated trajectories are reliable. Therefore, we apply a two-stage rejection sampling pipeline:
\begin{itemize}
    \item \textbf{Format validity filtering}: We discard trajectories that violate structural constraints, ensuring consistent CoT patterns.
    \item \textbf{Answer correctness filtering}: We filter out trajectories whose final answer does not match the ground-truth label in the training set.
\end{itemize}
In this way, we final curated set forms a high-quality CoT dataset for reasoning activation, and statistical details are provided in Appendix~\ref{Statistics of the CoT Dataset for SFT}. With this dataset in hand, given a question $Q$ and a filtered trajectory $y = [y_1, \dots, y_T]$, where $y$ concatenates the structured reasoning steps, tool calls, and final answer, the SFT objective maximizes the likelihood of the teacher trajectory:
\begin{equation}
\mathcal{L}_{\mathrm{SFT}} = 
- \sum_{t=1}^{T} \log \pi_{\theta}(y_t \mid Q, y_{<t}),
\label{eq:sft}
\end{equation}
where $\pi_\theta$ denotes the model’s token distribution. Following prior work~\cite{DBLP:journals/corr/abs-2508-13167}, we compute the loss only over model-generated tokens and exclude environment feedback (e.g., observations). The resulting model $\pi_{\mathrm{SFT}}$ serves as the initialization for the second online RL stage, substantially alleviating cold-start issues.

\subsection{Online RL with Temporal Tool Calls}
\label{Online RL with Temporal Tool Calls}

After the supervised fine-tuning stage, we further apply the online reinforcement learning with temporal tool calls, which enhances its multi-hop and time-sensitive reasoning capabilities. To enable effective learning in this setting, we address the problem from three key perspectives: action space, reward design, and training objective.

\subsubsection{Action Space}

As we can observe, existing retrievers are insensitive to temporal constraints and LLMs tend to overlook temporal requirements and hallucinate during question decomposition~\cite{PoK, guo2025empowering}, we identify the critical reasoning characteristics of temporal questions, and we categorize the reasoning paradigms of temporal reasoning over TKGs into specific taxonomies according to timestamps. Crucially, we formalize the agent’s action space as a collection of temporal functional tools. By doing so, we transform temporal reasoning into actionable primitives that directly mirror the TKG structure, moving beyond the limitations of the LLM’s implicit reasoning capabilities. In detail, our action space includes \textit{think}, \textit{plan}, temporal search actions, and \textit{answer}, with the temporal search actions defined as follows:
\begin{itemize}

    \item \textit{Search\_time(query)}. This action returns timestamps or time intervals associated with relevant quadruples.

    \item \textit{Search\_specific(query, $t$)}. This action returns relevant quadruples at the specified time $t$. 

    \item \textit{Search\_before(query, $t$)}. This action returns relevant quadruples occurring strictly before the specified time $t$. 

    \item \textit{Search\_after(query, $t$)}. This action returns relevant quadruples occurring strictly after the specified time $t$.

    \item \textit{Search\_between(query, $t_1, t_2$)}. This action returns relevant quadruples occurring within the time interval $[t_1, t_2]$.

\end{itemize}

Based on this action space and treating TKGs as the environment, temporal search tool returns structured feedback enclosed within \textit{<observation>} and \textit{</observation>}. Under this formulation, we adopt a ReAct-style~\cite{ReAct,DBLP:conf/www/LiuZLYPY25} interaction protocol, in which the reasoning process proceeds as a sequence of planning, internal thought generation, temporal search tool calls, and environment feedback. Formally, the interaction trajectory at step $n$ can be represented as:
\begin{equation}
\mathcal{H}_n = (\tau_0, p, \tau_1, a_1, o_1, \ldots, \tau_{n-1}, a_{n-1}, o_{n-1}),
\label{eq:tkg_trajectory}
\end{equation}
where $\tau$ denotes the agent's internal thought, $a$ is an action selected from the temporal search actions, with $p$ being the initial planning action, and $o$ is the observation obtained by executing the action $a$ over the TKGs. Based on the historical trajectory $\mathcal{H}_n$, the generation process for the next thought $\tau_n$ and action $a_n$ can be formulated as:
\begin{equation}
\pi_\theta(\tau_n \mid \mathcal{H}_n) = \prod_{i=1}^{|\tau_n|} \pi_\theta(\tau_n^i \mid \mathcal{H}_n, \tau_n^{<i}),
\label{eq:tkg_thought_policy}
\end{equation}
\begin{equation}
\pi_\theta(a_n \mid \mathcal{H}_n, \tau_n) = \prod_{j=1}^{|a_n|} \pi_\theta(a_n^j \mid \mathcal{H}_n, \tau_n, a_n^{<j}),
\label{eq:tkg_action_policy}
\end{equation}
where $\pi_\theta = \pi_{\mathrm{SFT}}$, $\tau_n^i$ and $|\tau_n|$ denote the $i$-th token and the length of $\tau_n$, and $a_n^j$ and $|a_n|$ denote the $j$-th token and the length of $a_n$. The interaction loop terminates when either the \textit{answer} action is invoked or the interaction-turn budget $B_{\max}$ is reached.

\subsubsection{Reward Design}
To optimize the above interactive process, we adopt the RLVR framework equipped with a novel multi-reward formulation. We incorporate three key components into our reward design: the \textit{format reward}, the \textit{retrieval reward}, and the \textit{outcome reward}.

\noindent \textbf{Format Reward} verifies whether the generated rollout adheres to the structured interaction protocol. Specifically, we ensure that the entire rollout follows the temporal interaction pattern defined in Eq.~\ref{eq:tkg_trajectory}. Formally, the format reward is defined as:
\begin{equation}
R_{\mathrm{fmt}} = \alpha\, \mathbb{I}_{\mathrm{fmt}},
\label{eq:format_reward}
\end{equation}
where $\mathbb{I}_{\mathrm{fmt}} \in \{0,1\}$ is a binary format validity indicator and $\alpha \in (0,1)$ is a scaling coefficient.

\noindent \textbf{Retrieval Reward} measures whether the retriever successfully retrieves evidence containing the correct answer, defined as:
\begin{equation}
R_{\mathrm{ret}} = \gamma\,\mathbb{I}_{\mathrm{ret}},
\label{eq:retrieval_reward}
\end{equation}
where $\mathbb{I}_{\mathrm{ret}} \in \{0,1\}$ is a binary retrieval indicator, and $\gamma\in (0,1)$ is a tunable scaling coefficient.

\noindent \textbf{Outcome Reward} evaluates the correctness of the final output answer $a_{\mathrm{pred}}$ by comparing it against the ground truth answer $a_{\mathrm{gold}}$ using a rule-based criteria exact match (EM):
\begin{equation}
R_{\mathrm{out}} = \mathrm{EM}(a_{\mathrm{pred}}, a_{\mathrm{gold}}),
\label{eq:outcome_reward}
\end{equation}
where $\mathrm{EM}(\cdot,\cdot)$ returns $1$ if the two strings match exactly and $0$ otherwise, and thus $R_{\mathrm{out}} \in \{0,1\}$. Finally, we combine the above components into the overall reward. The final reward $R_{\mathrm{all}}$ is defined as:

\begin{equation}
\begin{aligned}
R_{\mathrm{all}}
= {} & R_{\mathrm{out}}\!\left(1 - (1-\mathbb{I}_{\mathrm{fmt}})\lambda\right)
\\[4pt]
&+ (1 - R_{\mathrm{out}})\!\left(R_{\mathrm{fmt}} + R_{\mathrm{ret}}\right)
\\[4pt]
&+ (1 - R_{\mathrm{out}})\,\delta\left(1-\mathbb{I}_{\mathrm{fmt}}\right),
\end{aligned}
\label{eq:final_reward}
\end{equation}
where $\lambda > 0$ denotes the penalty applied when the answer is correct but the format is invalid, and $\delta > 0$ serves as a fallback reward granted when both the answer and format are incorrect. In this way, $R_{\mathrm{fmt}}$ enforces adherence to the temporal interaction protocol, $R_{\mathrm{ret}}$ promotes effective problem decomposition and evidence retrieval over TKGs, and $R_{\mathrm{out}}$ ensures factual correctness.

\subsubsection{Training Objective} With the multi-dimensional reward formulation defined, we formalize the overall training objective of \OurMODEL{} framework. To ensure robust policy optimization, we adopt the RLVR paradigm, which can be instantiated through either PPO or GRPO. 

The policy $\pi_\theta$ is optimized by maximizing the objective function $\mathcal{J}(\theta)$, which encourages trajectories with higher-than-average rewards while maintaining stability via importance sampling and Kullback–Leibler (KL) divergence constraints. The overall objective is defined as:
\begin{equation}
\begin{aligned}
\mathcal{J}(\theta) = & \hat{\mathbb{E}}_{Q, {y_i}_{i=1}^G \sim \pi_{\text{old}}} \left[ \frac{1}{G} \sum_{i=1}^G f_\epsilon(\rho_i(\theta), \hat{A_i}) \right] \\ & - \beta \cdot \hat{\mathbb{E}}_{Q} \left[ \mathbb{D}_{KL} \left[ \pi_\theta(\cdot | Q) || \pi_{\text{ref}}(\cdot | Q) \right] \right],
\end{aligned}
\end{equation}
where $\rho_i(\theta) = \frac{\pi_\theta(y_i | Q)}{\pi_{\text{old}}(y_i | Q)}$ denotes the importance sampling ratio for the $i$-th trajectory, and $f_\epsilon(\rho_i(\theta), \hat{A}) = \min(\rho_i(\theta) \hat{A}, \text{clip}(\rho_i(\theta), 1 - \epsilon, 1 + \epsilon)\hat{A})$ is the clipping function. For standard PPO, $G=1$ and the advantage $\hat{A}_i$ is estimated using a learned value function, whereas for GRPO, $G>1$ and the advantage $\hat{A}_i$ is computed group-relatively.

\section{Experiments}
In this section, we evaluate \OurMODEL{} on widely used datasets. We conduct extensive experiments to demonstrate the effectiveness of our method by answering the following research questions (RQ): (1) \textbf{RQ1:} How does \OurMODEL{} perform compared to state-of-the-art baselines on complex TKGQA datasets?  (2) \textbf{RQ2:} What is the contribution of each key module in the \OurMODEL{} framework to the overall performance? (3) \textbf{RQ3:} How do different retrieval configurations affect reasoning performance? (4) \textbf{RQ4:} How does RL optimization shape the model's behavior in TKGQA scenarios? We also conduct a cross-domain generalization study in Appendix~\ref{sec:Cross-domain Generalization Study}, and present a case study in Appendix~\ref{sec:Case Study} to further demonstrate the robustness and the advantages of our proposed method.

\subsection{Experimental Settings}
\label{sec: Experimental Settings}

We evaluate \OurMODEL{} on representative TKGQA benchmarks, including MULTITQ~\cite{MULTITQ} and CronQuestions~\cite{CronKGQA}. Detailed descriptions of these datasets are provided in Appendix~\ref{Details of Dataset}, while additional results on the cross-domain benchmark TimelineKGQA~\cite{DBLP:conf/www/SunLH0L25} are reported in Appendix~\ref{sec:Cross-domain Generalization Study}. We adopt Hits@1 as the evaluation metric, measuring the proportion of questions for which the top-ranked prediction is correct. We compare \OurMODEL{} against three categories of baselines: PLM-based methods, Embedding-based methods, and LLM-based methods. A detailed description of all baseline models is provided in Appendix~\ref{Baselines}. For training, we use GPT-4o as the teacher model to generate trajectories and adopt e5-base-v2~\cite{e5-base-v2} for evidence retrieval, retrieving the top-15 most relevant quadruples per query. \OurMODEL{} is instantiated with Llama3-8B-Instruct, Qwen2.5-7B-Instruct, and Qwen3-4B-Instruct-2507 as backbone models. Additional implementation details are provided in Appendix~\ref{Implementation Details}.

\begin{table*}[htb]
\centering
\small
\setlength{\tabcolsep}{3.5pt}
\begin{tabular}{c|c cc cc|c cc cc}
\toprule
& \multicolumn{5}{c|}{\textbf{MULTITQ}}
& \multicolumn{5}{c}{\textbf{CronQuestions}} \\
\cmidrule(lr){2-6}\cmidrule(lr){7-11}
\multirow{2}{*}{\textbf{Methods}}
& \multirow{2}{*}{\textbf{Overall}} 
& \multicolumn{2}{c}{\textbf{Question Type}}
& \multicolumn{2}{c|}{\textbf{Answer Type}}
& \multirow{2}{*}{\textbf{Overall}} 
& \multicolumn{2}{c}{\textbf{Question Type}}
& \multicolumn{2}{c}{\textbf{Answer Type}} \\
\cmidrule(lr){3-4}\cmidrule(lr){5-6}\cmidrule(lr){8-9}\cmidrule(lr){10-11}
& 
& \textbf{Single} & \textbf{Multiple}
& \textbf{Entity} & \textbf{Time}
& 
& \textbf{Simple} & \textbf{Complex}
& \textbf{Entity} & \textbf{Time} \\
\midrule

\rowcolor{gray!15} \multicolumn{11}{c}{\textbf{\textit{PLM-based Methods}}} \\
\midrule

BERT~\cite{BERT}     & 0.083 & 0.092 & 0.061 & 0.101 & 0.040 & 0.243 & 0.249 & 0.239 & 0.277 & 0.179 \\
DistillBERT~\cite{DistilBERT} & 0.083 & 0.087 & 0.074 & 0.102 & 0.037 & -- & -- & -- & -- & -- \\
ALBERT~\cite{ALBERT}      & 0.108 & 0.116 & 0.086 & 0.139 & 0.032 & 0.248 & 0.255 & 0.235 & 0.279 & 0.177 \\
\midrule

\rowcolor{gray!15} \multicolumn{11}{c}{\textbf{\textit{Embedding-based Methods}}} \\
\midrule

EmbedKGQA~\cite{EmbedKGQA} & 0.206 & 0.235 & 0.134 & 0.290 & 0.001 & 0.288 & 0.290 & 0.286 & 0.411 & 0.057 \\
CronKGQA~\cite{CronKGQA}  & 0.279 & 0.337 & 0.134 & 0.328 & 0.156 & 0.647 & 0.987 & 0.392 & 0.699 & 0.549 \\
MultiQA~\cite{MULTITQ}   & 0.293 & 0.347 & 0.159 & 0.349 & 0.157 & -- & -- & -- & -- & -- \\
\midrule

\rowcolor{gray!15} \multicolumn{11}{c}{\textbf{\textit{LLM-based Methods}}} \\
\midrule

ARI~\cite{ARI}       & 0.380 & 0.680 & 0.210 & 0.394 & 0.344 & 0.707 & 0.860 & 0.570 & 0.660 & 0.800 \\
Naive RAG~\cite{NaiveRAG} & 0.379 & 0.469 & 0.155 & 0.242 & 0.672 & 0.633 & 0.726 & 0.280 & 0.610 & 0.684 \\
ReAct RAG~\cite{ReAct} & 0.398 & 0.506 & 0.130 & 0.243 & 0.735 & 0.809 & 0.863 & 0.600 & 0.768 & 0.895 \\
TempAgent~\cite{TempAgent} & 0.702 & 0.857 & 0.316 & 0.624 & 0.870 & 0.842 & 0.895 & 0.640 & 0.805 & 0.921 \\
TimeR$^{4}$~\cite{TimeR4} & 0.728 & 0.887 & 0.335 & 0.639 & 0.945 & -- & -- & -- & -- & -- \\
RTQA~\cite{gong-etal-2025-rtqa}  & 0.765 & 0.902 & 0.424 & 0.692 & 0.942 & -- & -- & -- & -- & -- \\
PoK~\cite{PoK}   & 0.779 & \textbf{0.929} & 0.409 & 0.696 & \underline{0.962} & -- & -- & -- & -- & -- \\
\midrule

\rowcolor{gray!15} \multicolumn{11}{c}{\textbf{\textit{\OurMODEL{} (Ours)}}} \\
\midrule

\ding{168}~~Qwen2.5-7B-Instruct & \textbf{0.855} & 0.910 & \textbf{0.721} & \textbf{0.814} & 0.955  & \underline{0.893} & \underline{0.958} & \underline{0.844} & \underline{0.863} & \underline{0.949} \\ 
\ding{171}~~Qwen2.5-7B-Instruct & \underline{0.824} & 0.881 & \underline{0.683} & \underline{0.774} & 0.945 & 0.868 & 0.937 & 0.817 & 0.838 & 0.925 \\

\cmidrule(lr){1-11}

\ding{168}~~Llama3-8B-Instruct  & 0.810 & 0.872 & 0.659 & 0.756 & 0.943 & \textbf{0.915} & \textbf{0.968} & \textbf{0.875} & \textbf{0.893} & \textbf{0.955} \\
\ding{171}~~Llama3-8B-Instruct  & 0.799 & 0.861 & 0.644 & 0.709 & 0.739 & 0.873 & 0.949 & 0.817 & 0.844 & 0.928 \\

\cmidrule(lr){1-11}

\ding{168} Qwen3-4B-Instruct  & 0.804 & \underline{0.914} & 0.533 & 0.739 & \textbf{0.963} & 0.889 & 0.953 & 0.837 & 0.857 & 0.948 \\
\ding{171}~~Qwen3-4B-Instruct  & 0.780 & 0.901 & 0.480 & 0.708 & 0.956 & 0.877 & 0.953 & 0.820 & 0.841 & 0.944 \\

\bottomrule
\end{tabular}

\caption{Performance comparison of baselines and \OurMODEL{} in Hits@1 across different question and answer types on MULTITQ and CronQuestions. \ding{168} denotes \OurMODEL{} trained with SFT+GRPO, while \ding{171} denotes training with SFT+PPO. The best and second-best scores are marked in \textbf{bold} and \underline{underline}, respectively.}

\label{tab: Main results}
\end{table*}

\subsection{Main Results (RQ1)}
\label{sec: Main results}

In this section, we compare \OurMODEL{} with representative baselines on MULTITQ and CronQuestions. As shown in Table~\ref{tab: Main results}, \OurMODEL{} achieves consistently superior overall performance, outperforming diverse baselines across different model families, parameter scales (4B, 7B, 8B), and RL training strategies (trained with GRPO or PPO), demonstrating strong model-agnostic applicability. Compared to the strongest baseline, \OurMODEL{} achieves absolute overall Hits@1 improvements of $7.60\%$ and $7.30\%$ on MULTITQ and CronQuestions, respectively. These results suggest that enabling LLMs to dynamically interact with TKGs via RAG mechanisms facilitates effective search strategies and temporally grounded reasoning capabilities. Notably, \OurMODEL{} exhibits substantial improvements on complex multi-step TKGQA tasks, surpassing the best-performing baselines on the corresponding complex settings by $29.70\%$ on MULTITQ (Multiple) and $23.50\%$ on CronQuestions (Complex). This further confirms that our approach significantly enhances temporal multi-hop reasoning through explicit planning and time-aware retrieval tool usage. While \OurMODEL{} shows slightly lower performance on Single-type questions in MULTITQ, this difference can be reasonably attributed to PoK’s use of a retriever specifically optimized for single-step temporal retrieval.

\begin{table*}[!htbp]
\centering
\small
\setlength{\tabcolsep}{2.8pt}
\begin{tabular}{l|c|cc|cccccc}
\toprule
\multirow{2}{*}{\textbf{Model}} 
& \multirow{2}{*}{\textbf{Overall}} 
& \multicolumn{2}{c|}{\textbf{Answer Type}} 
& \multicolumn{6}{c}{\textbf{Fine-grained Question Type}} \\
\cmidrule(lr){3-4} \cmidrule(lr){5-10}
& & \textbf{Entity} & \textbf{Time}
& \textbf{Equal} & \textbf{Before/After} & \textbf{First/Last} 
& \textbf{Equal Multi} & \textbf{After First} & \textbf{Before Last} \\
\midrule

\textbf{TKG-Thinker} & \textbf{0.855} & \textbf{0.814} & \underline{0.955} & \textbf{0.939} & \textbf{0.851} & \textbf{0.920} & \textbf{0.792} & \textbf{0.678} & \textbf{0.729} \\

\textit{w/o SFT Stage} & 0.591 & 0.465 & 0.897 & 0.869 & 0.609 & 0.836 & 0.213 & 0.078 & 0.084 \\

\textit{w/o Plan Action} & \underline{0.796} & \underline{0.766} & 0.871 & \underline{0.933} & \underline{0.822} & 0.838
& \underline{0.704} & \underline{0.538} & \underline{0.609} \\

\textit{w/o Temporal Retrievers} & 0.458 & 0.252 & \textbf{0.959} & 0.466 & 0.456 & \underline{0.909} & 0.205 & 0.133 & 0.139 \\
\bottomrule
\end{tabular}
\caption{Ablation study on the MULTITQ dataset. \textbf{Bold} indicates the best performance, while \underline{underline} marks the second-best . Single-type questions include Equal, Before/After, and First/Last; Multiple-type questions include Equal Multi, After First, and Before Last. “\textit{w/o}” means removing or replacing the corresponding module.}
\label{tab:ablation_multitq}
\end{table*}

\begin{figure*}[t]
    \centering
    \subfigure{
        \includegraphics[width=0.48\textwidth]{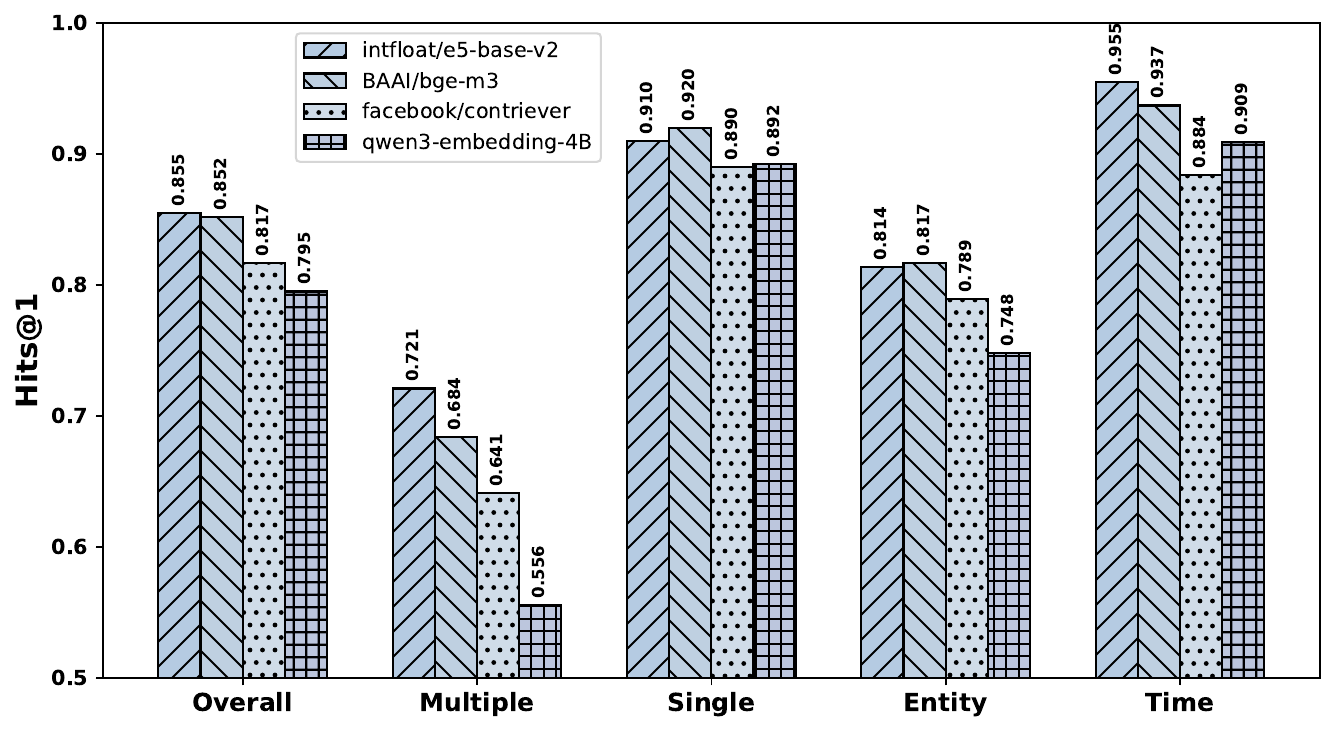}
        \label{fig:retriever_comparison_MULTITQ}
    }
    \hfill
    \subfigure{
        \includegraphics[width=0.48\textwidth]{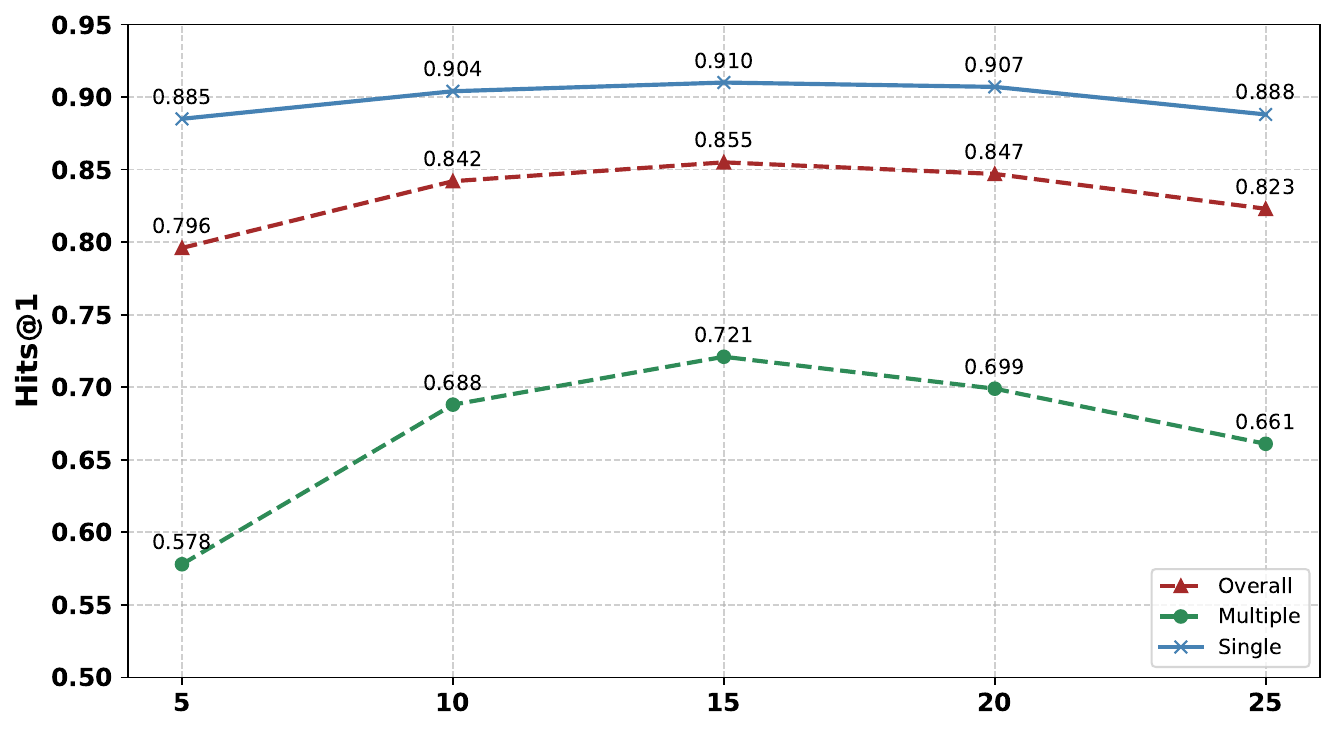}
        \label{fig:retrieval_topk_MULTITQ}
    }
    \caption{Retriever analysis on the MULTITQ dataset. \textbf{Left}: Performance comparison of different retriever models. \textbf{Right}: Effect of retrieval depth, measured by the number of top-$k$ retrieved quadruples.}
    \label{fig:retriever_analysis_MULTITQ}
\end{figure*}

\begin{figure*}[t]
    \centering
    \includegraphics[width=\textwidth]{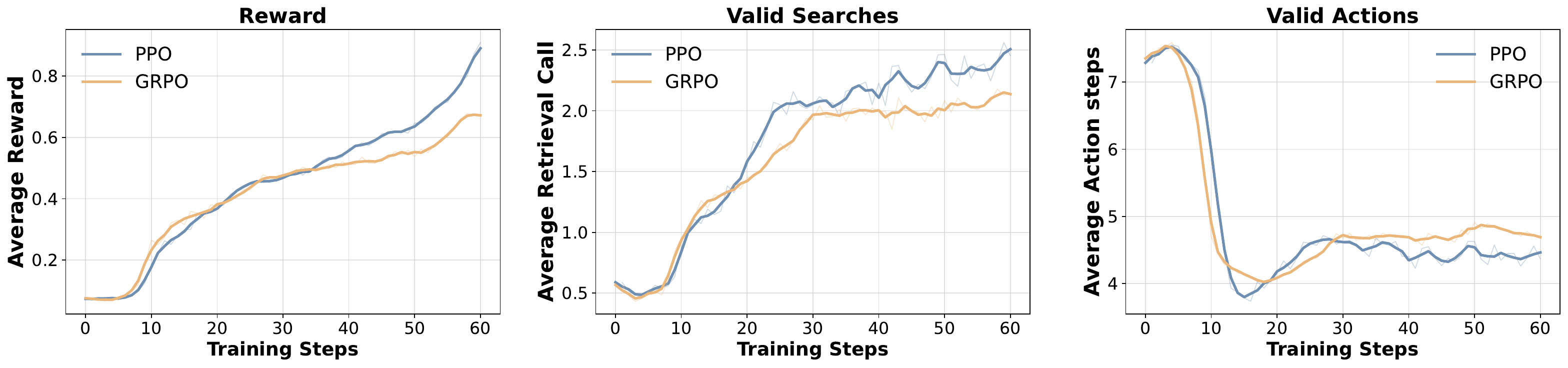}
    \caption{Training dynamics of \OurMODEL{} implemented with GRPO and PPO on MULTITQ. \textbf{Left}: Training Reward; \textbf{Middle}: Retrieval Call Steps; \textbf{Right}: Action Steps.}
    \label{fig:ppo_grpo_reward_search_action_compare}
\end{figure*}

\subsection{Ablation Study (RQ2)}
\label{sec: Ablation Study}

In this section, we conduct a series of ablation experiments to examine the contribution of each component in \OurMODEL{}, as summarized in Table~\ref{tab:ablation_multitq}, including the SFT stage, the planning mechanism, and the temporal retrievers. Specifically, we systematically remove or replace individual components to construct corresponding model variants for comparison. Additional ablation results on the CronQuestions are reported in Appendix~\ref{sec:Ablation Study on CronQuestion}.

\noindent \textbf{Effect of the SFT Stage.} We further analyze the role of SFT, which initializes the model’s ability to execute structured reasoning protocols prior to reinforcement learning. When the SFT stage is removed and the model is trained with RL alone, the overall performance drops drastically by $26.40\%$. This confirms that SFT provides essential scaffolding that stabilizes learning, reduces temporal hallucination, enables verifiable temporal reasoning rather than unconstrained free-form generation.

\noindent \textbf{Effect of the Plan Action.} We remove the Plan action and prompt the model to interact directly with the environment using the original queries. As a result, eliminating the planning component leads to an overall performance drop of 5.90\%. In particular, performance on Multiple-type temporal questions decreases by 8.80\%, 14.00\%, and 12.00\% on Equal Multi, After First, and Before Last, respectively. These results suggest that planning component plays an important role in reliable temporal reasoning over TKGs, as removing it tends to  hallucinated intermediate reasoning steps.

\noindent \textbf{Effect of the Temporal Retrievers.} To assess the role of temporal retrievers, we replace them with a purely semantic retriever that ignores temporal constraints. This substitution yields the largest performance degradation (\textminus $39.70\%$), highlighting the importance of temporal alignment between questions and evidence. Notably, performance drops sharply on multiple-type temporal questions, demonstrating that temporal retrieval is indispensable for providing fine-grained, temporally grounded evidence to support reliable temporal reasoning.

\subsection{Retrieval Analysis (RQ3)}
\label{sec: Retrieval Analysis}

\noindent \textbf{Effect of Retriever Model.}  Since retrieval quality directly determines the availability of temporal evidence, we first evaluate \OurMODEL{} under four representative retrievers: \texttt{e5-base-v2}, \texttt{bge-m3}, \texttt{contriever}, and \texttt{qwen3-embedding-4B}. As shown in Figure~\ref{fig:retriever_analysis_MULTITQ} (Left), all four retrievers achieve competitive performance, substantially surpassing the strongest baseline (PoK). Among them, contrastively trained retrievers (e.g., \texttt{e5-base-v2}, \texttt{bge-m3}) deliver superior performance, with the advantage being more evident on Multiple questions.

\noindent \textbf{Effect of Retrieval Depth.}  The hyperparameter $k$ controls how many top-ranked quadruples the temporal search tools return as environmental feedback. As illustrated in Figure~\ref{fig:retriever_analysis_MULTITQ} (Right), performance increases with $k$ and then declines. This reflects a trade-off: larger $k$ improves the likelihood of retrieving useful evidence, whereas excessively large $k$ introduces distractors that impede LLM reasoning. Notably, performance degradation at larger $k$ is more pronounced on Multiple questions. We attribute this phenomenon to the accumulation of errors across successive reasoning steps, as Multiple-type questions require iterative retrieval and multi-step reasoning, where early errors are progressively amplified in later stages. In practice, we find that $k=15$ offers the best balance between evidence coverage and distractor noise.

\subsection{Training Dynamics (RQ4)}
\label{sec: Training Dynamics}
To investigate how \OurMODEL{} evolves during the training process, we illustrate its training dynamics in Figure~\ref{fig:ppo_grpo_reward_search_action_compare} (with additional entropy and response details in Appendix~\ref{sec:Training Dynamics Details}). As shown in the left panel, both PPO and GRPO exhibit a steady increase in training rewards. This demonstrates that our fine-grained reward design provides stable reinforcement signals, facilitating consistent policy optimization.
Regarding action and retrieval dynamics, we observe a clear "decrease–then–increase" pattern. Specifically, the average number of action steps initially drops sharply as the model learns to follow the required output format and eliminates redundant or invalid actions. Subsequently, both action steps and retrieval calls gradually increase and stabilize, indicating that \OurMODEL{} strategically invokes additional temporal tool calls to acquire necessary evidence and thereby strengthens its agentic reasoning capability. Notably, while both algorithms converge well, PPO achieves a higher reward ceiling and more frequent retrieval calls in the later stages of training. 

\section{Conclusion}

we introduce \OurMODEL{}, a novel agent equipped with autonomous planning and adaptive retrieval capabilities for reasoning over TKGs. By modeling the TKG as a dynamic environment, \OurMODEL{} integrates supervised fine-tuning and reinforcement learning with a multi-reward optimization scheme to enhance temporal reasoning. Experiments show that \OurMODEL{} consistently outperforms baselines, demonstrating the effectiveness of explicit interaction and RL-driven optimization in reducing hallucination and improving multi-step reasoning. 

\section*{Limitations}
Despite \OurMODEL{}'s strong performance achieved in TKGQA, this work has several limitations. The current reward mechanism relies heavily on binary indicators and rule-based criteria, such as Exact Match (EM) for outcomes and basic format verification. This outcome-based reward lacks a nuanced evaluation of the intermediate reasoning process. Future iterations could incorporate an LLM Judge with detailed rubrics to qualitatively assess the logical and temporal consistency of the think and plan steps, ensuring that the model understands complex temporal constraints rather than just optimizing for a specific output format. Besides, while the model demonstrates effective multi-step reasoning, the relative simplicity of current datasets and benchmarks—particularly their limited reasoning hops—restricts the training of temporal agents capable of long-range planning and inference. Future work should thus explore more complex synthetic multi-hop tasks and open-world settings to foster greater model robustness.

\section*{Ethics Statement}

In constructing the CoT-based SFT datasets, we have taken into account ethical considerations and limitations commonly associated with large language models. All data used in this work are publicly available and do not contain personal or sensitive information. Nonetheless, we acknowledge that, despite our best efforts, the datasets may still contain gaps or unintended biases. To mitigate these concerns, the source data has been curated to ensure diversity and reduce potential bias. Through careful dataset construction, review, and testing procedures, we strive to uphold ethical AI principles while advancing research in TKGQA.


\bibliography{custom}


\appendix

\section{Experimental Settings}
\label{sec:Experimental Settings}

\subsection{Few-shot Prompt for Initial Trajectory Generation}
\label{Few-shot Prompt for Initial Trajectory Generation}

\begin{figure*}[!t]
    \centering
    \includegraphics[width=\linewidth]{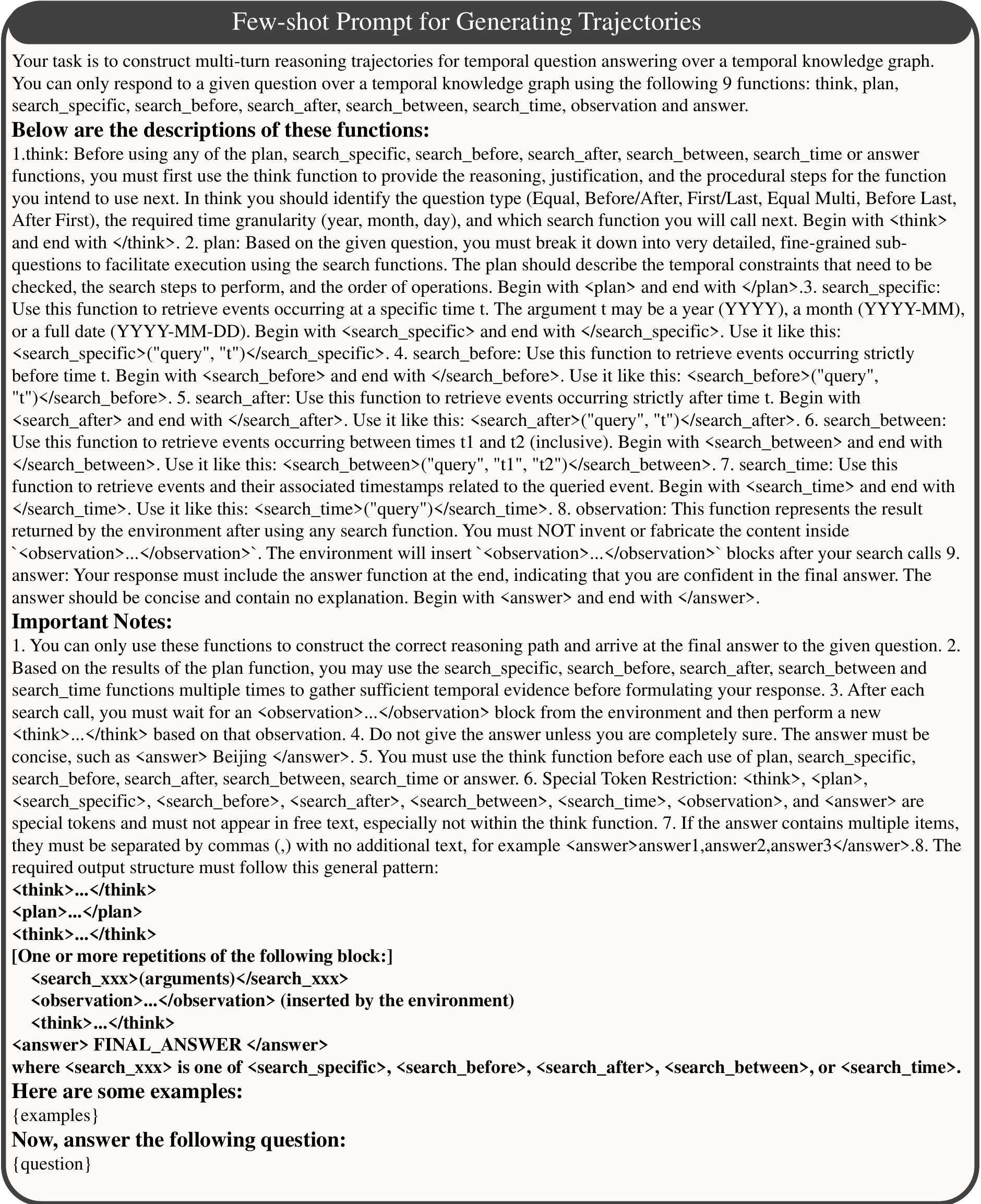}
    \caption{Few-shot Prompt for Generating Trajectories.}
    \label{fig:Few-shot Prompt for Generating Trajectories}
\end{figure*}

As shown in Figure~\ref{fig:Few-shot Prompt for Generating Trajectories}, we adopt a few-shot prompting strategy to elicit structured, tool-augmented reasoning during trajectory generation. This stage provides the initial pool of trajectories before subsequent format and answer-correctness filtering.

\subsection{Dataset Details}
\label{Details of Dataset}

\begin{table}[ht]
\centering
\small
\begin{tabular}{l l r r r}
\toprule
\textbf{Type} & \textbf{} & \textbf{Train} & \textbf{Valid} & \textbf{Test} \\
\midrule
\multirow{3}{*}{Single} 
    & Equal         & 135,890 & 18,983 & 17,311 \\
    & Before/After  & 75,340  & 11,655 & 11,073 \\
    & First/Last    & 72,252  & 11,097 & 10,480 \\
\midrule
\multirow{3}{*}{Multiple}
    & Equal Multi   & 16,893 & 3,213 & 3,207 \\
    & After First   & 43,305 & 6,499 & 6,266 \\
    & Before Last   & 43,107 & 6,532 & 6,247 \\
\midrule
\textbf{Total} & & \textbf{386,787} & \textbf{57,979} & \textbf{54,584} \\
\bottomrule
\end{tabular}
\caption{Data statistics of MULTITQ.}
\label{tab:multitq}
\end{table}

\begin{table}[htb]
\centering
\small
\begin{tabular}{lccc}
\toprule
\textbf{Type} & \textbf{Train} & \textbf{Valid} & \textbf{Test} \\
\midrule
Simple Entity & 90,651 & 7,745 & 7,812 \\
Simple Time & 61,471 & 5,197 & 5,046 \\
Before/After & 23,869 & 1,982 & 2,151 \\
First/Last & 118,556 & 11,198 & 11,159 \\
Time Join & 55,453 & 3,878 & 3,832 \\
\midrule
Simple Reasoning & 152,122 & 12,942 & 12,858 \\
Complex Reasoning & 197,878 & 17,058 & 17,142 \\
\midrule
Entity Answer & 225,672 & 19,362 & 19,524 \\
Time Answer & 124,328 & 10,638 & 10,476 \\
\midrule
\textbf{Total} & \textbf{350,000} & \textbf{30,000} & \textbf{30,000} \\
\bottomrule
\end{tabular}
\caption{Dataset Statistics of CronQuestions.}
\label{tab:cron_questions}
\end{table}

\paragraph{MULTITQ.} MULTITQ is a large-scale temporal question answering dataset that incorporates multi-granularity temporal information. It provides a comprehensive evaluation protocol across several dimensions: Question Type (Multiple vs. Single), Answer Type (Entity vs. Time), and Time Granularity (year, month, and day). The detailed statistics of MULTITQ are presented in Table~\ref{tab:multitq}.

\paragraph{CronQuestions.} 
CronQuestions is a temporal QA benchmark consisting of 410K unique question–answer pairs. Its questions can be categorized into two major types: Simple temporal reasoning (e.g., Simple Entity and Simple Time) and Complex temporal reasoning (e.g., Before/After, First/Last, and Time-Join queries), depending on the temporal constraints involved. Detailed statistics of CronQuestions are shown in Table~\ref{tab:cron_questions}.

\subsection{Statistics of the CoT Dataset for SFT}
\label{Statistics of the CoT Dataset for SFT}

\begin{table}[htb]
\centering
\small
\caption{Statistics of the SFT datasets constructed from MULTITQ and CronQuestions.}
\resizebox{0.9\linewidth}{!}{
\begin{tabular}{l|l|r}
\toprule
\textbf{Dataset} & \textbf{Question Type} & \textbf{Count} \\
\midrule
\multirow{7}{*}{MULTITQ}
& Equal            & 4017 \\
& Before/After     & 1668 \\
& First/Last       & 1062 \\
& Equal Multi      & 1395 \\
& After First      & 1074 \\
& Before Last      & 1055 \\

\cmidrule(lr){2-3}
& \textbf{Total}   & \textbf{10271} \\
\midrule
\multirow{6}{*}{CronQuestions}
& Simple Entity     & 2509 \\
& Simple Time      & 1726 \\
& Before/After    & 668 \\
& First/Last      & 3301 \\
& Time Join         & 1508 \\
\cmidrule(lr){2-3}
& \textbf{Total}   & \textbf{9712} \\
\bottomrule
\end{tabular}
}
\label{tab:sft_data_stats}
\end{table}

To mitigate the cold-start issue in reinforcement learning, we construct CoT-style supervised fine-tuning datasets from MULTITQ and CronQuestions. As summarized in Table~\ref{tab:sft_data_stats}, these datasets cover diverse temporal reasoning types and provide explicit supervision signals for temporal decomposition and retrieval behaviors.

\subsection{Baseline Details}
\label{Baselines}

we compare \OurMODEL{} against three categories of baselines: (1) \textbf{PLM-based methods}, including BERT\cite{BERT}, ALBERT\cite{ALBERT}, and DistilBERT\cite{DistilBERT}; (2) \textbf{Embedding-based methods}, such as EmbedKGQA\cite{EmbedKGQA}, CronKGQA\cite{CronKGQA}, and MultiQA\cite{MULTITQ}; and (3) \textbf{LLM-based methods}, including  Naive RAG~\cite{NaiveRAG}, ReAct RAG\cite{ReAct}, ARI\cite{ARI}, RTQA\cite{gong-etal-2025-rtqa}, and TempAgent\cite{TempAgent}, TimeR4~\cite{TimeR4}, and PoK~\cite{PoK}. For consistency with prior work, we adopt the baseline results reported in \citet{PoK}, \citet{ARI}, and \citet{TempAgent} for comparison.

\subsection{Implementation Details}
\label{Implementation Details}

\begin{table}[htb]
\centering
\caption{Key hyperparameters used in model training.}
\resizebox{\linewidth}{!}{
\begin{tabular}{c|c}
\toprule
\textbf{Category} & \textbf{Value} \\
\midrule
\multicolumn{2}{l}{\textit{Supervised Fine-tuning (SFT)}} \\
\midrule
Finetuning Type & Full \\
Epochs & 4 \\
Batch Size & 4 \\
Grad. Accumulation & 4 \\
Learning Rate & $1\times10^{-5}$ \\
Scheduler & Cosine Decay \\
Warmup Ratio & 0.03 \\
Cutoff Length & 8192 \\
Precision & BF16 \\
\midrule
\multicolumn{2}{l}{\textit{Reinforcement Learning (RL)}} \\
\midrule
Algorithm & PPO / GRPO \\
Training Steps & 80 \\
Train Batch Size & 256 \\
Max Turns ($B_{\max}$) & 8 \\
Temperature & 0.7 \\
Retriever Top-$k$ & 15 \\
Max Response Length (per turn) & $512$ \\
Max Observation Length (per turn) & $1024$ \\
Format Reward Coef. ($\alpha$) & $0.2$ \\
Outcome Penalty Coef. ($\lambda$) & $0.4$ \\
Retrieval Reward Coef. ($\gamma$) & $0.1$ \\
Fallback Reward Coef. ($\delta$) & $0.1$ \\
\midrule
\multicolumn{2}{l}{\textit{Inference}} \\
\midrule
Temperature & $0.01$ \\
Top-$p$ & $0.95$ \\
Max New Tokens & $512$ \\
Max Infer Step & $8$ \\
\bottomrule
\end{tabular}
}
\label{tab:train_hyper}
\end{table}

As summarized in Table~\ref{tab:train_hyper}, during the SFT stage, we fine-tune the models via the LLaMA-Factory framework~\cite{zheng2024llamafactory} with a batch size of~4 for~4 epochs using AdamW with a learning rate of~$1\mathrm{e}{-5}$ and cosine decay scheduling. In the RL stage, we switch to the Verl framework and train both PPO and GRPO policies with a batch size of~256, a mini-batch size of~32, and~5 rollouts. We set the interaction-turn budget to $B_{\max}=8$. For rollout collection, we apply sampling with a temperature of~0.7 and conduct retrieval-augmented interactions using the top-$15$ evidence candidates per query. To stabilize trajectory generation and prevent excessive growth of reasoning tokens, model-generated responses and retrieved observations are truncated to $512$ and $1024$ tokens per turn, respectively. Regarding the training data, the SFT stage uses rejection sampling to retain trajectories, while the RL stage is trained on TKGQA data not used in SFT (5,001 QA pairs for MULTITQ and 5,230 for CronQuestions). During inference, we disable stochastic sampling and adopt deterministic decoding with temperature~$0.01$, and top-$p=0.95$. All experiments are implemented in PyTorch and conducted on 8 NVIDIA A800 (80GB) GPUs.


\section{Cross-domain Generalization Study}
\label{sec:Cross-domain Generalization Study}

\begin{table}[htb]
\centering
\resizebox{\columnwidth}{!}{
\begin{tabular}{l l r r r}
\hline
\textbf{Dataset} & \textbf{Type} & \textbf{Train} & \textbf{Val} & \textbf{Test} \\
\hline
\multirow{4}{*}{Timeline-CronQuestion}
& Simple  & 7,200  & 2,400 & 2,400 \\
& Medium  & 8,252  & 2,751 & 2,751 \\
& Complex & 9,580  & 3,193 & 3,193 \\
\cline{2-5}
& \textbf{Total}   & \textbf{25,032} & \textbf{8,344} & \textbf{8,344} \\
\hline
\multirow{4}{*}{Timeline-ICEWS}
& Simple  & 17,982 & 5,994 & 5,994 \\
& Medium  & 15,990 & 5,330 & 5,330 \\
& Complex & 19,652 & 6,550 & 6,550 \\
\cline{2-5}
& \textbf{Total}   & \textbf{53,624} & \textbf{17,874} & \textbf{17,874} \\
\hline
\end{tabular}
}
\caption{Data statistics of Timeline-CronQuestion and Timeline-ICEWS.}
\label{tab:timeline}
\end{table}

\begin{table}[htbp]
\centering
\small
\resizebox{\linewidth}{!}{
\begin{tabular}{l c c c c}
\toprule
\textbf{Model} & \textbf{Overall} & \textbf{Simple} & \textbf{Medium} &\textbf{Complex}\\
\midrule
RAG baseline      & 0.235 & 0.704 & 0.092 & 0.009  \\
LLaMA2-7B & 0.169 & 0.049 & 0.143 & 0.282 \\
GPT-4o    & 0.206 & 0.069 & 0.130 & 0.376 \\
RTQA    & 0.298 & \underline{0.608} & 0.218 & 0.135 \\
\midrule
\rowcolor{gray!15} \multicolumn{5}{c}{\textbf{\textit{\OurMODEL{} (Ours)}}} \\
\midrule
Qwen2.5-7B-Instruct  & \underline{0.460} & 0.567 & \underline{0.294} & \underline{0.523} \\
Llama3-8B-Instruct  & \textbf{0.491} & \textbf{0.612} & \textbf{0.320} & \textbf{0.546} \\
\bottomrule
\end{tabular}
}
\caption{Results on the Timeline-CronQuestion dataset across different reasoning difficulty levels. \textbf{Bold} indicates the best performance and \underline{underline} denotes the second best performance.}
\label{tab:Results on the Timeline-CronQuestion}
\end{table}

\begin{table}[htbp]
\centering
\small
\resizebox{\linewidth}{!}{
\begin{tabular}{l c c c c}
\toprule
\textbf{Model} & \textbf{Overall} & \textbf{Simple} & \textbf{Medium} &\textbf{Complex}\\
\midrule
RAG baseline      & 0.265 & 0.660 & 0.128 & 0.011  \\
LLaMA2-7B & 0.111 & 0.035 & 0.066 & 0.322 \\
GPT-4o    & 0.113 & 0.051 & 0.035 & 0.353 \\
\midrule
\rowcolor{gray!15} \multicolumn{5}{c}{\textbf{\textit{\OurMODEL{} (Ours)}}} \\
\midrule
Qwen2.5-7B-Instruct  & \textbf{0.508} & \underline{0.556} & \textbf{0.409} & \textbf{0.583} \\
Llama3-8B-Instruct  & \underline{0.462} & \textbf{0.596} & \underline{0.273} & \underline{0.533} \\
\bottomrule
\end{tabular}
}
\caption{Results on the Timeline-ICEWS dataset across different reasoning difficulty levels. \textbf{Bold} indicates the best performance, while \underline{underline} denotes the second-best performance.}
\label{tab:Results on the Timeline-ICEWS}
\end{table}

\begin{table*}[!htbp]
\centering
\small
\setlength{\tabcolsep}{2.8pt}
\begin{tabular}{l|c|cc|ccccc}
\toprule
\multirow{2}{*}{\textbf{Model}} 
& \multirow{2}{*}{\textbf{Overall}} 
& \multicolumn{2}{c|}{\textbf{Answer Type}} 
& \multicolumn{5}{c}{\textbf{Fine-grained Question Type}} \\
\cmidrule(lr){3-4} \cmidrule(lr){5-9}
& & \textbf{Entity} & \textbf{Time}
& \textbf{Simple Entity} & \textbf{Simple Time} & \textbf{Before/After} & \textbf{First/Last} 
& \textbf{Time Join}  \\
\midrule

\textbf{TKG-Thinker} & \textbf{0.893} & \textbf{0.863} & \underline{0.949} & \textbf{0.946} & \textbf{0.977} & \textbf{0.699} & \underline{0.891} & \textbf{0.787} \\

\textit{w/o SFT Stage} & 0.638 & 0.732 & 0.568 & 0.861 & 0.532 & 0.487 & 0.564 & 0.623 \\

\textit{w/o Plan Action} & \underline{0.860} & \underline{0.811} & \textbf{0.951} & \underline{0.942} & \textbf{0.977} & \underline{0.538} & \textbf{0.899} & 0.603 \\

\textit{w/o Temporal Retrievers} & 0.821 & 0.756 & 0.942 & 0.924 & \underline{0.963} & 0.477 & \underline{0.891} & 0.414 \\
\bottomrule
\end{tabular}
\caption{Ablation study on the CronQuestions dataset. \textbf{Bold} indicates the best performance, while \underline{underline} marks the second-best. Simple-type questions include Simple Entity, and Simple Time; Complex-type questions include Before/After, First/Last, and Time Join. “\textit{w/o}” means removing or replacing the corresponding module.}

\label{tab:ablation_cron}
\end{table*}

We further evaluate the cross-domain generalizability of \OurMODEL{}. Specifically, we evaluate \OurMODEL{} on the TimelineKGQA~\cite{{DBLP:conf/www/SunLH0L25}} benchmark, which includes Timeline-CronQuestions and Timeline-ICEWS. These datasets differ in both temporal representations and the complexity of temporal reasoning. In TimelineKGQA, Simple questions require a single contextual fact and typically involve temporally constrained retrieval or timeline position identification; Medium questions require two contextual facts and further involve the combination of retrieval with temporal semantic operations and timeline arithmetic; Complex questions require three contextual facts and cover the full spectrum of temporal reasoning capabilities. The dataset statistics are summarized in Table~\ref{tab:timeline}. \textbf{For this evaluation, we use Qwen2.5-7B-Instruct and Llama3-8B-Instruct as backbones, both trained on MULTITQ using SFT and GRPO.} Following~\citet{gong-etal-2025-rtqa} and~\citet{DBLP:conf/www/SunLH0L25}, we compare \OurMODEL{} with a RAG baseline, LLaMA2-7B, GPT-4o, and RTQA. For consistency with prior work, baseline results are sourced from~\citet{gong-etal-2025-rtqa}.

As shown in Table~\ref{tab:Results on the Timeline-CronQuestion} and Table~\ref{tab:Results on the Timeline-ICEWS}, \OurMODEL{} consistently outperforms all baselines in Overall score, and the results reveal three consistent trends. First, \OurMODEL{} exhibits clear performance gains as temporal reasoning complexity increases. While improvements on Simple questions are marginal, the gains on Medium and Complex questions are substantially larger, indicating that \OurMODEL{} is particularly effective at handling compositional temporal reasoning and timeline arithmetic. Second, the relative advantage of \OurMODEL{} is preserved across datasets with distinct temporal characteristics: on Timeline-CronQuestion the improvements are most salient on Medium and Complex queries, whereas on Timeline-ICEWS the model maintains strong performance across all difficulty levels despite its larger event space and more diverse temporal expressions. Third, LLaMA2-7B and GPT-4o exhibit limited ability in settings that require explicit temporal grounding, whereas retrieval-augmented methods that rely on sub-question decomposition, such as RTQA, partially close the performance gap but still struggle with multi-hop temporal composition. These findings suggest that structured tool-augmented trajectories and agentic decision-making are key to improving temporal reasoning capabilities in various TKGQA scenarios.

\begin{figure*}[htb]
    \centering
    \includegraphics[width=\textwidth]{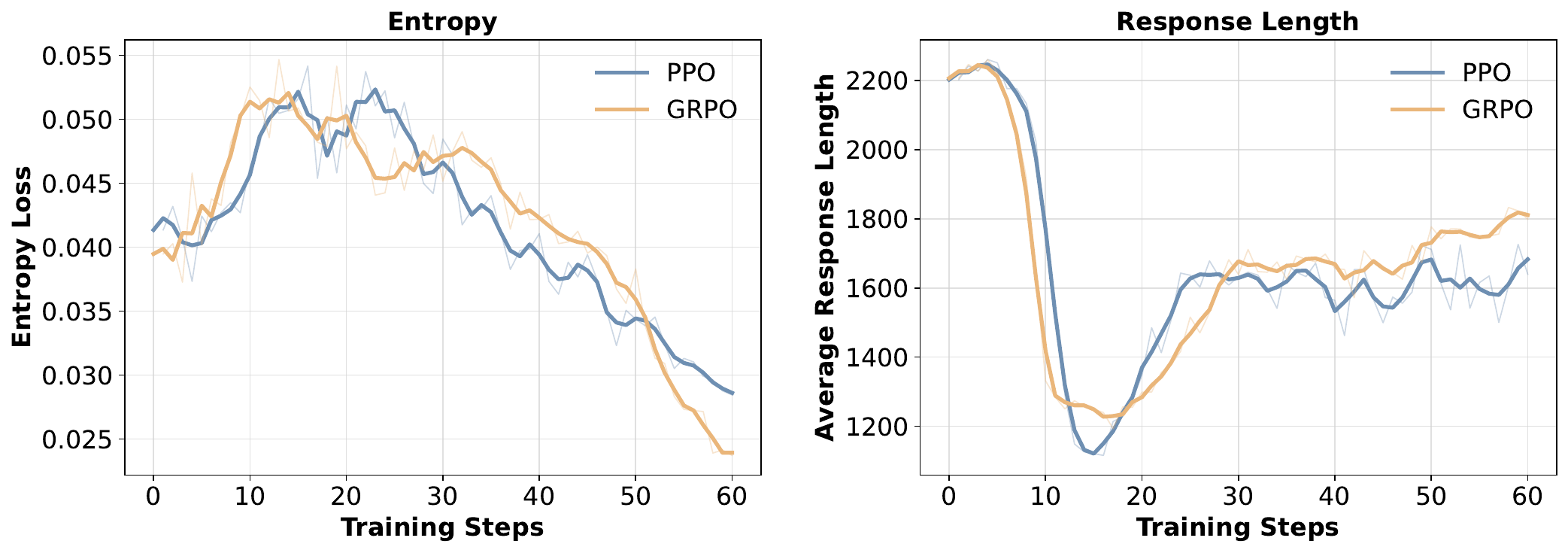}
    \caption{Training dynamics of \OurMODEL{} implemented with GRPO and PPO on MULTITQ. \textbf{Left}: Generation Entropy; \textbf{Right}: Response Length.}
    \label{fig:ppo_grpo_entropy_response_compare}
\end{figure*}

\section{Ablation Study on CronQuestion}
\label{sec:Ablation Study on CronQuestion}

We further conduct ablation experiments on CronQuestions to analyze the contribution of each component in \OurMODEL{}, as summarized in Table~\ref{tab:ablation_cron}. Consistent with the findings on MULTITQ, several clear observations can be drawn. First, removing the SFT stage and training the model with reinforcement learning alone leads to a substantial performance drop of 25.50\% in overall accuracy. This result indicates that SFT is essential for alleviating the cold-start problem in reinforcement learning by providing a stable initialization and structured reasoning priors. Second, eliminating the Plan action leads to a noticeable performance decline of 3.30\% in the overall score, with the most severe degradation observed on Complex-type temporal questions. This suggests that explicit planning is particularly important for handling complex temporal dependencies that require multi-step reasoning. Finally, replacing the temporal retrievers with a purely semantic retriever that disregards temporal constraints leads to the performance drop of $7.2\%$. This result highlights that explicit temporal alignment between the question and the retrieved evidence is a fundamental prerequisite for accurate TKGQA. Overall, these results demonstrate that \OurMODEL{} benefits from the combined effects of supervised initialization, explicit planning, and temporally aware retrieval to achieve robust temporal reasoning.

\section{Training Dynamics Details}
\label{sec:Training Dynamics Details}
Figure~\ref{fig:ppo_grpo_entropy_response_compare} illustrates the evolution of training entropy and response length, providing deeper insights into the training process of \OurMODEL{}.

\paragraph{Training Entropy.} The left panel shows that after an initial exploration phase, the policy entropy for both PPO and GRPO consistently declines. This trend indicates that the model is successfully converging from diverse exploration to a stable, optimized policy by exploiting high-reward reasoning paths.

\paragraph{Response Length.} The right panel exhibits a distinct "V-shaped" trajectory: 
\begin{itemize}
    \item \textbf{Alignment Phase (Steps 0-15):} A sharp decline in length occurs as the model learns to prune redundant tokens and strictly follow the required agentic interaction format.
    \item \textbf{Reasoning Expansion (Steps 15-60):} The length gradually increases and stabilizes, suggesting that once the format is mastered, \OurMODEL{} learns to generate more substantial reasoning chains and necessary tool calls to solve complex temporal questions. 
\end{itemize}

\begin{figure*}[!t]
    \centering
    \includegraphics[width=\textwidth]{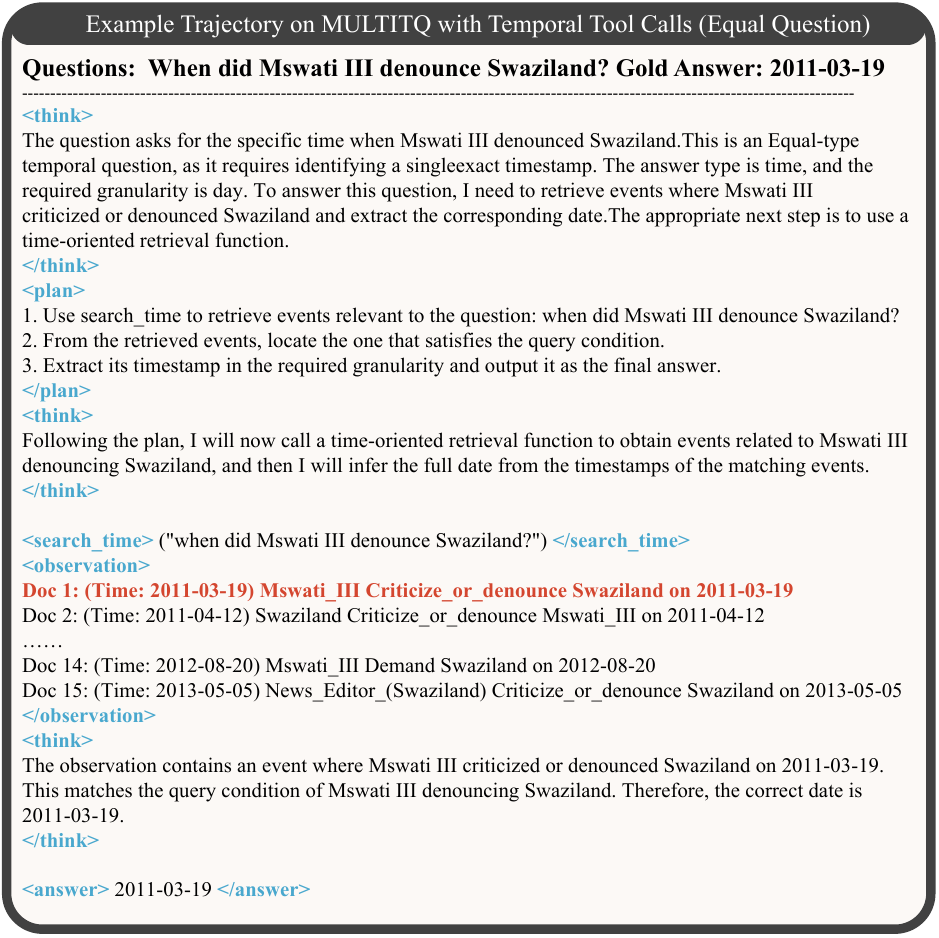}
    \caption{A Case Study on MULTITQ for Equal Questions.}
    \label{fig:Case_Study_multitq_1}
\end{figure*}

\begin{figure*}[!t]
    \centering
    \includegraphics[width=\textwidth]{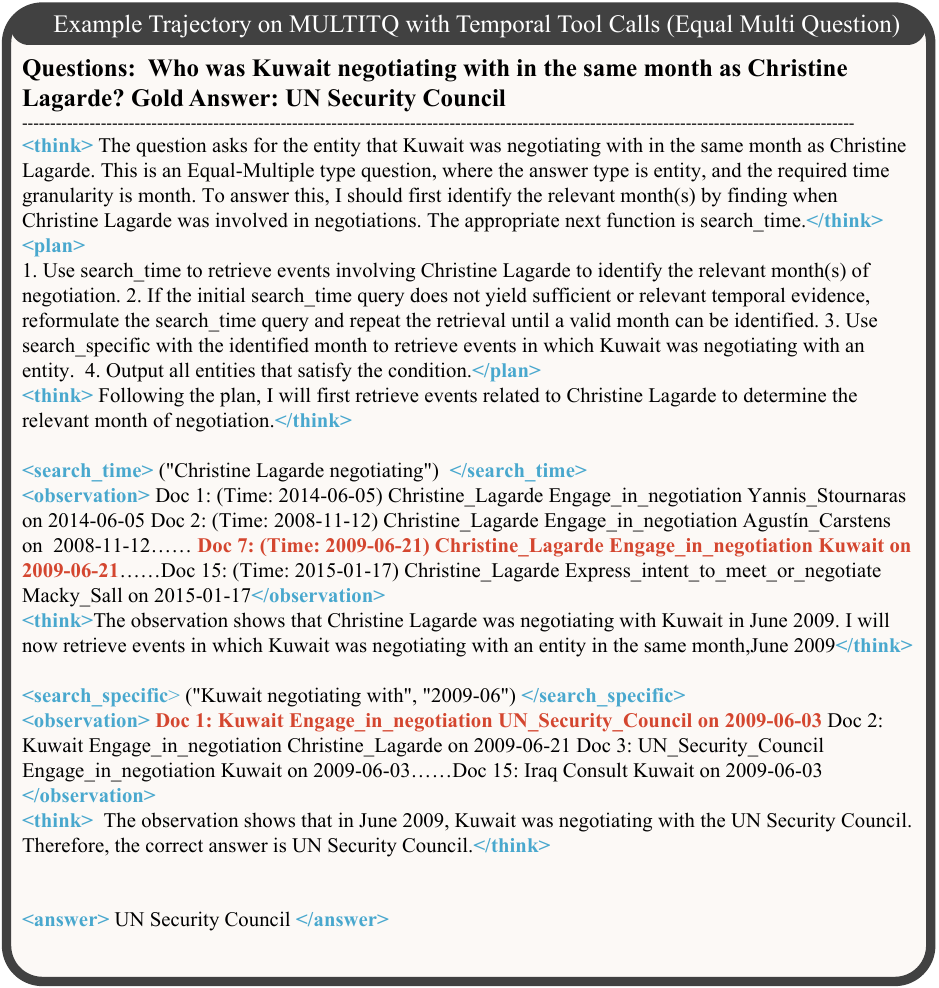}
    \caption{A Case Study on MULTITQ for Equal Multi Questions.}
    \label{fig:Case_Study_multitq_2}
\end{figure*}

\begin{figure*}[!t]
    \centering
    \includegraphics[width=\textwidth]{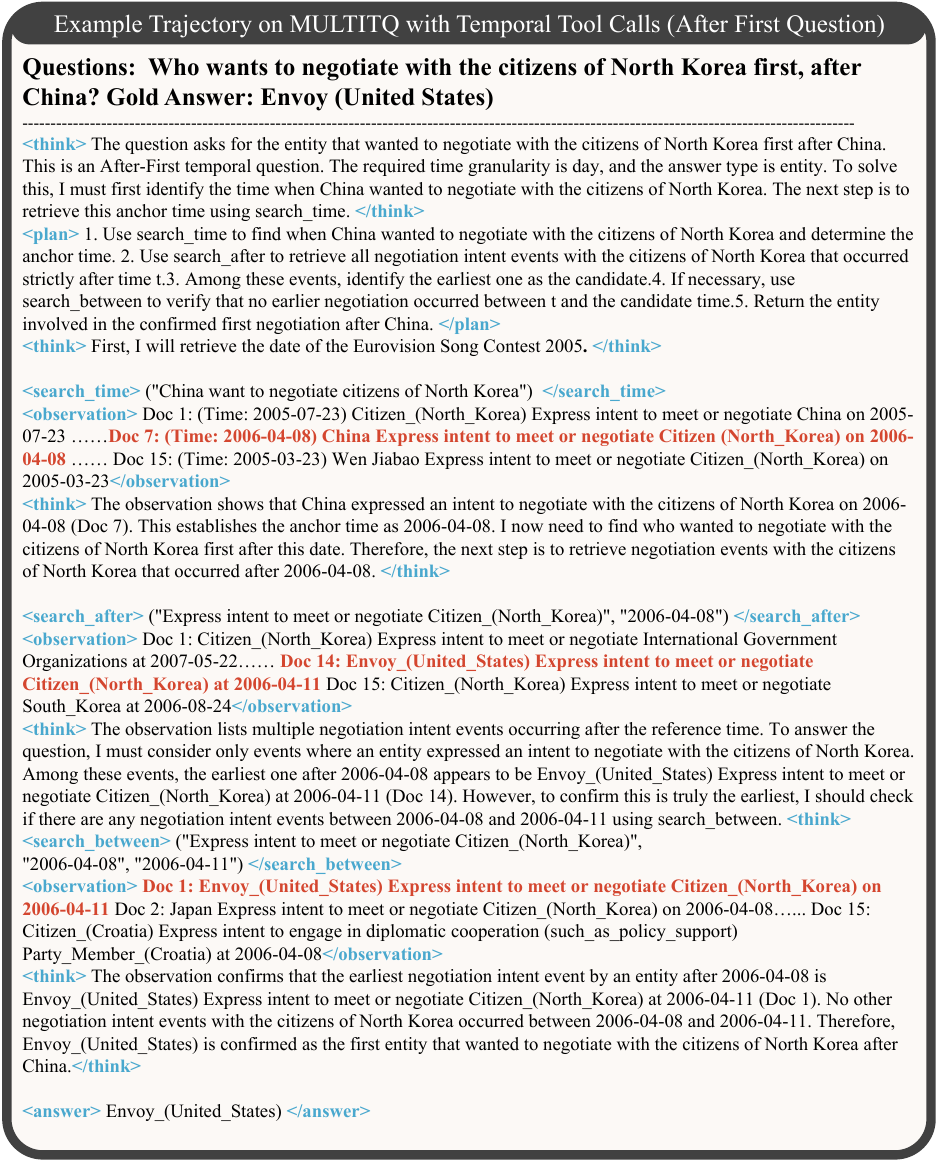}
    \caption{A Case Study on MULTITQ for After First Questions.}
    \label{fig:Case_Study_multitq_3}
\end{figure*}

\begin{figure*}[!t]
    \centering
    \includegraphics[width=\textwidth]{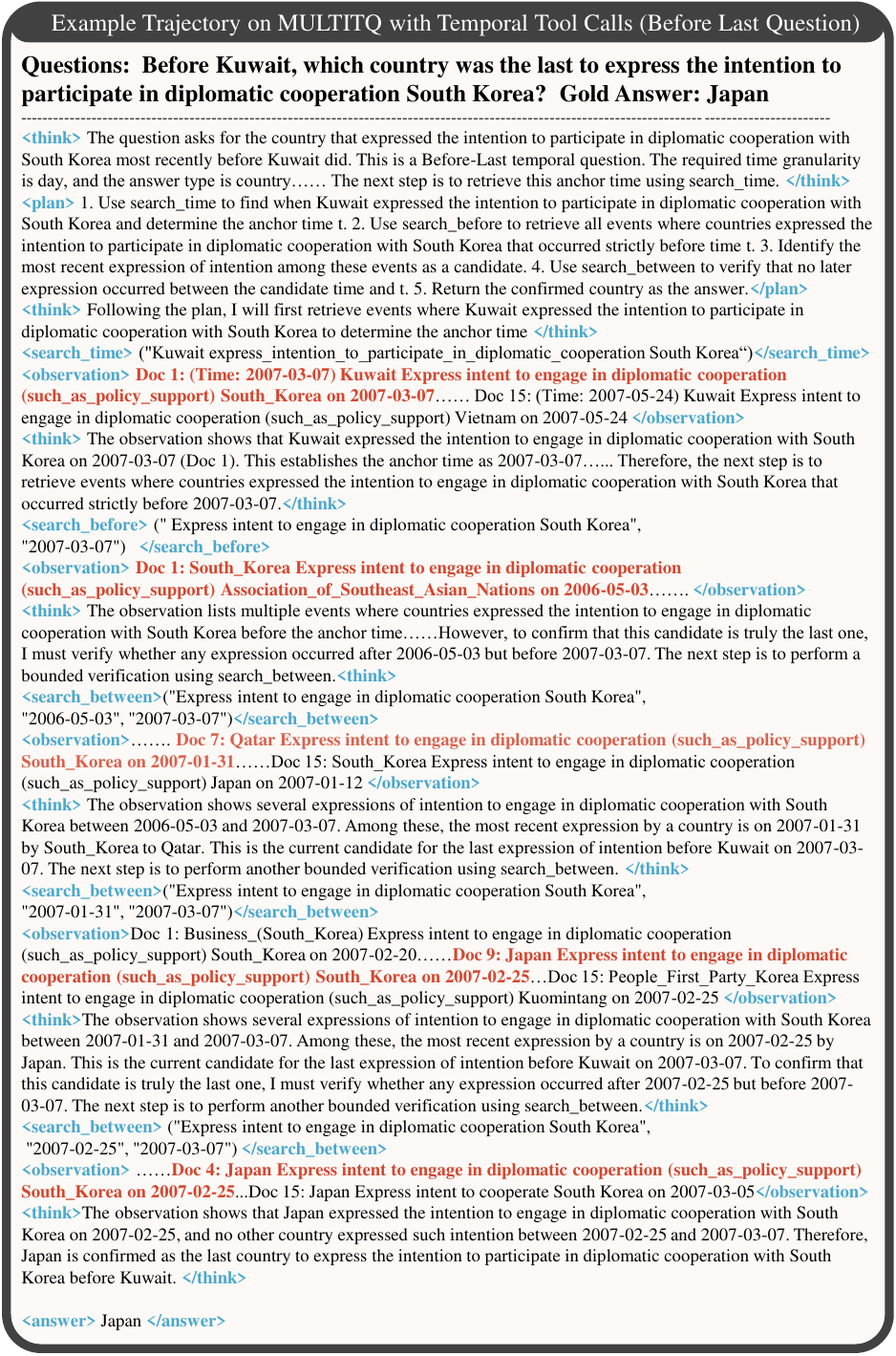}
    \caption{A Case Study on MULTITQ for Before Last Questions.}
    \label{fig:Case_Study_multitq_4}
\end{figure*}

\section{Case Study}
\label{sec:Case Study}

To illustrate how \OurMODEL{}, equipped with autonomous planning and adaptive retrieval, performs temporal reasoning over TKGs to obtain correct answers, we analyze several representative questions. 

\paragraph{MULTITQ.} Figures~\ref{fig:Case_Study_multitq_1} and~\ref{fig:Case_Study_multitq_2} present case studies of Equal and Equal Multi questions, respectively. These examples demonstrate that \OurMODEL{} can accurately identify relevant temporal conditions, retrieves supporting quadruples for each constraint, and integrates the evidence to derive the final answer. Figures~\ref{fig:Case_Study_multitq_3} and~\ref{fig:Case_Study_multitq_4} further illustrate After First and Before Last questions. In these cases, \OurMODEL{} first plans an anchor event, then performs iterative retrieval under strict temporal constraints, and crucially conducts internal temporal verification to ensure that no earlier or later events violate the query requirements. For example, Figure~\ref{fig:Case_Study_multitq_4} shows that \OurMODEL{} performs bounded verification via repeated \textit{Search\_between(query, $t_1, t_2$)} calls, updating the candidate from Association\_of\_Southeast\_Asian\_Nations to Qatar and finally to Japan. These examples provide concrete evidence of the model’s verification behavior during temporal reasoning.


\begin{figure*}[!t]
    \centering
    \includegraphics[width=\textwidth]{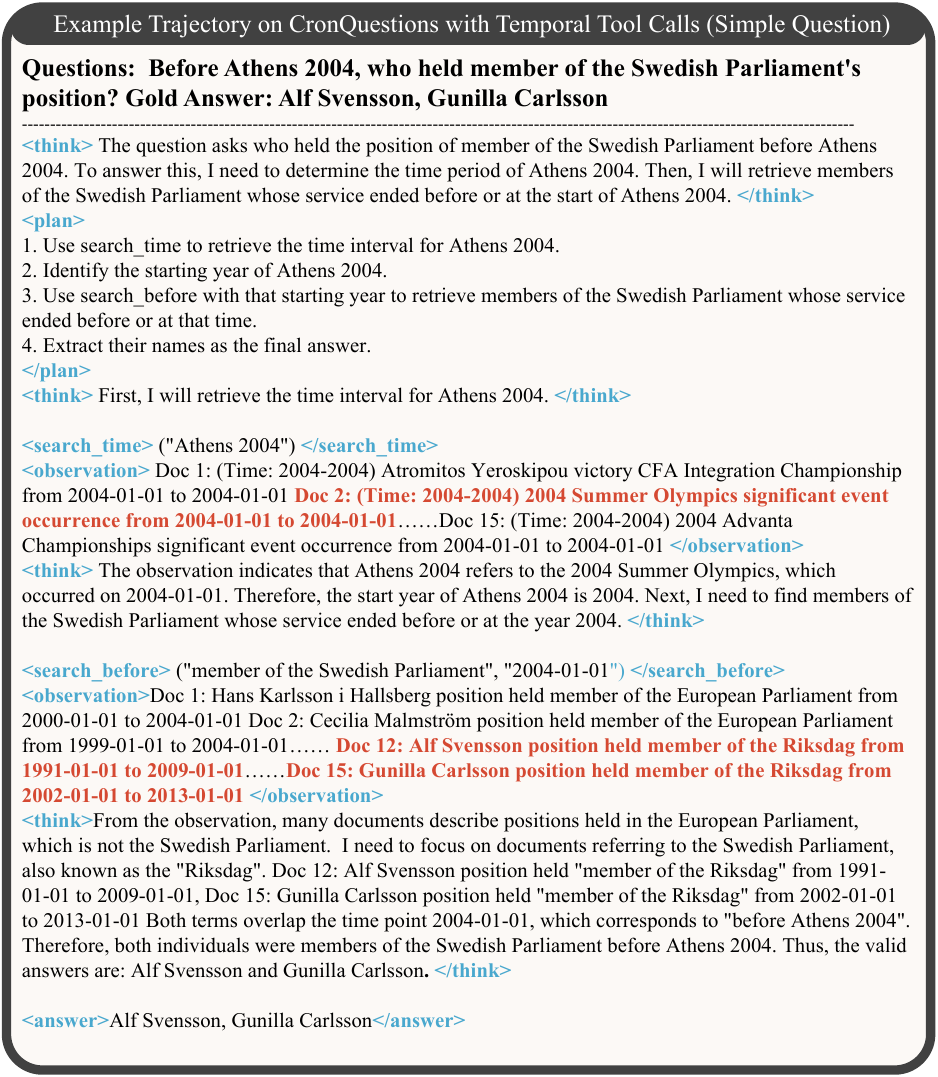}
    \caption{A Case Study on CronQuestions for Simple TKGQA Questions.}
    \label{fig:Case_Study_cron_1}
\end{figure*}

\begin{figure*}[!t]
    \centering
    \includegraphics[width=\textwidth]{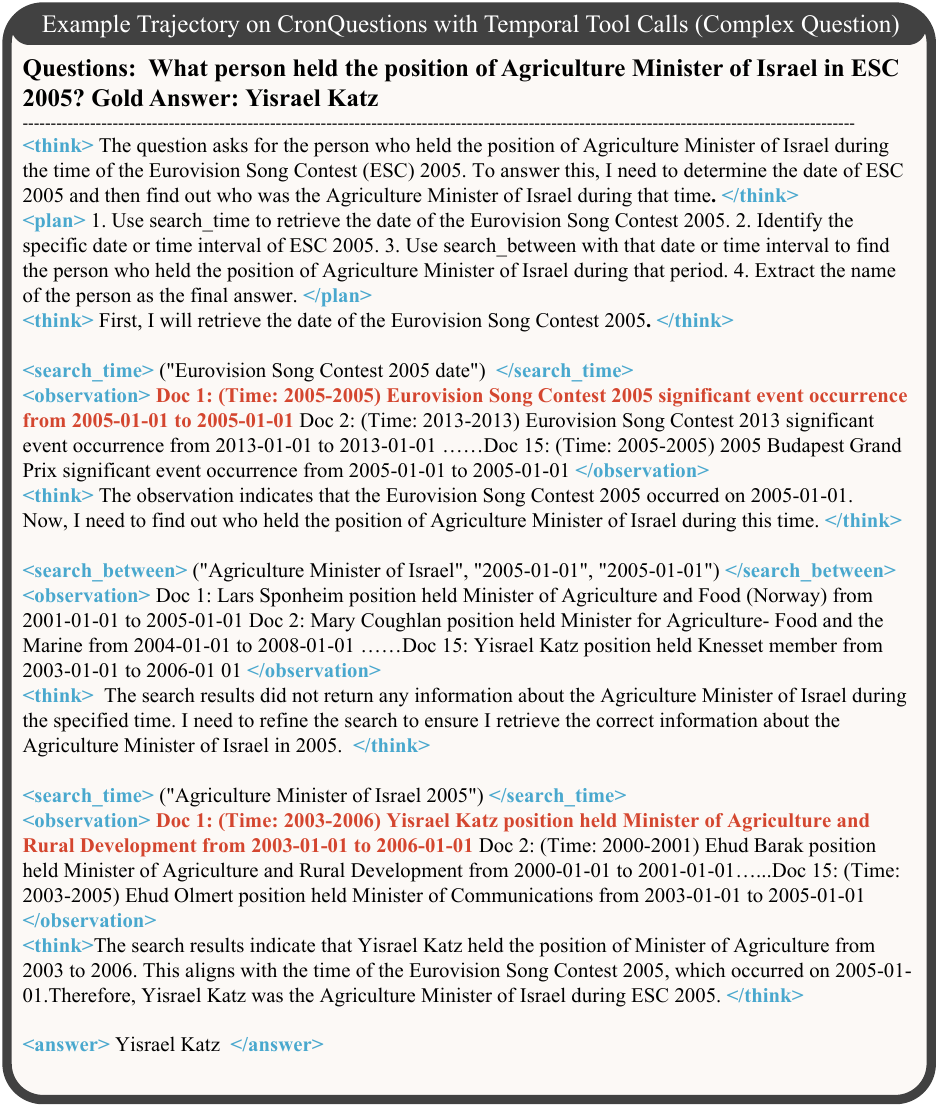}
    \caption{A Case Study on CronQuestions for Complex TKGQA Questions.}
    \label{fig:Case_Study_cron_2}
\end{figure*}

\paragraph{CronQuestions.} Unlike MULTITQ, which represents facts as timestamped quadruples, CronQuestions models knowledge using quintuples with explicit temporal intervals. As illustrated in Figure~\ref{fig:Case_Study_cron_1} and Figure~\ref{fig:Case_Study_cron_2}, both simple and complex queries in CronQuestions require reasoning over interval-level temporal constraints. In this setting, \OurMODEL{} is able to decompose a given question into a sequence of executable subtasks via the plan action, perform iterative and temporally aligned retrieval, and ultimately produce the correct answer.

\begin{figure*}[!t]
    \centering
    \includegraphics[width=\textwidth]{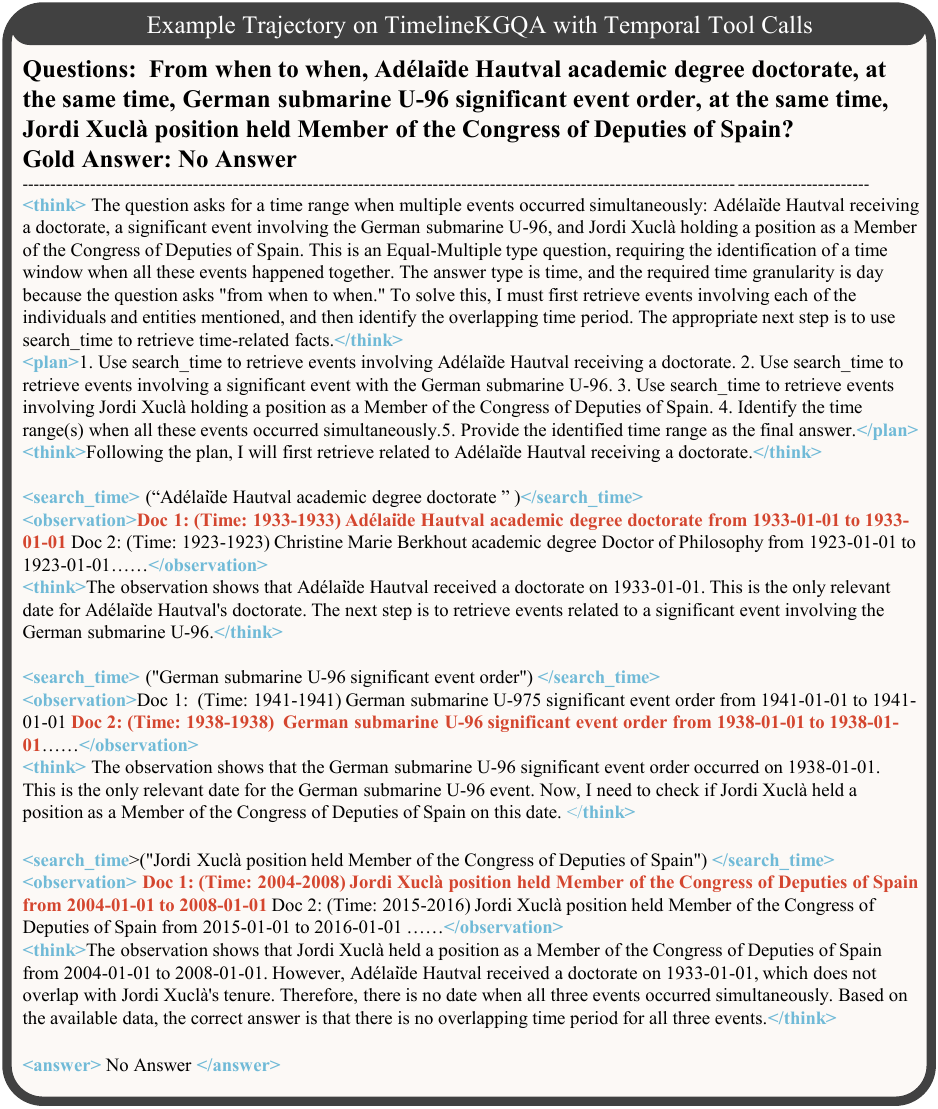}
    \caption{A Case Study on TimelineKGQA.}
    \label{fig:Case_Study_TimelineKGQA}
\end{figure*}

\paragraph{TimelineKGQA.} Figure~\ref{fig:Case_Study_TimelineKGQA} illustrates a TimelineKGQA example. Although \OurMODEL{} is not trained on this dataset, it successfully adapts its tool usage strategy, verifies temporal consistency, and produces the correct answer. This result demonstrates that \OurMODEL{} generalizes well to previously unseen temporal settings.

\end{document}